\documentclass[acmlarge, nonacm]{acmart}

\usepackage{multirow}
\usepackage{rotating}
\usepackage{url}

\usepackage{graphicx}
\usepackage{array}

\usepackage{adjustbox}
\usepackage{forest}

\newcommand*\rot{\rotatebox[origin=c]{45}}

\def\signed #1 (#2){{\unskip\nobreak\hfil\penalty50\hskip2em\hbox{}\nobreak\hfil\sl#1\/ \rm(#2)\parfillskip=0pt \finalhyphendemerits=0 \par}}

\AtBeginDocument{%
  }

\setcopyright{acmcopyright}
\copyrightyear{2022}
\acmYear{2022}
\acmDOI{XXXXXXX.XXXXXXX}

\acmJournal{CSUR}
\acmVolume{X}
\acmNumber{X}
\acmArticle{1}
\acmMonth{10}

\begin{document}

\title{Machine Generated Text: A Comprehensive Survey of Threat Models and Detection Methods}

\author{Evan Crothers}
\email{ecrot027@uottawa.ca}
\orcid{0000-0001-6177-0525}

\author{Nathalie Japkowicz}
\email{japkowic@american.edu}
\orcid{0000-0003-1176-1617}

\author{Herna Viktor}
\email{hviktor@uottawa.ca}
\orcid{0000-0003-1914-5077}

\renewcommand{\shortauthors}{Crothers et al.}

\begin{abstract}
Machine generated text is increasingly difficult to distinguish from human authored text.  Powerful open-source models are freely available, and user-friendly tools that democratize access to generative models are proliferating. ChatGPT, which was released shortly after the first edition of this survey, epitomizes these trends. The great potential of state-of-the-art natural language generation (NLG) systems is tempered by the multitude of avenues for abuse.  Detection of machine generated text is a key countermeasure for reducing abuse of NLG models, with significant technical challenges and numerous open problems.  We provide a survey that includes both 1) an extensive analysis of threat models posed by contemporary NLG systems, and 2) the most complete review of machine generated text detection methods to date.  This survey places machine generated text within its cybersecurity and social context, and provides strong guidance for future work addressing the most critical threat models, and ensuring detection systems themselves demonstrate trustworthiness through fairness, robustness, and accountability.
\end{abstract}

\ccsdesc[500]{Computing methodologies~Machine learning approaches}
\ccsdesc[500]{Security and privacy~Human and societal aspects of security and privacy}
\ccsdesc[300]{Computing methodologies~Neural networks}
\ccsdesc[500]{Computing methodologies~Natural language generation}

\keywords{machine learning, artificial intelligence, neural networks, trustworthy AI, machine generated text, transformer, text generation, threat modeling, cybersecurity, disinformation, generative AI}

\maketitle


\section{Introduction}

\subsection{Risks of Machine Generated Text}
\label{ssec:risks}

Recent natural language generation (NLG) models have taken a significant step forward in diversity, control, and quality of machine generated text.  The ability to create unique, manipulable, human-like text with unprecedented speed and efficiency presents additional technical challenges for the detection of abuses of NLG models, such as phishing \cite{10.1145/3052973.3053037, 10.1007/978-3-030-14687-0_8}, disinformation  \cite{shu2020mining, stiff2022detecting, zellers2019defending}, fraudulent product reviews \cite{10.1007/978-3-030-44041-1_114, stiff2022detecting}, academic dishonesty \cite{hargrave_2005, dehouche2021plagiarism}, and toxic spam \cite{kurenkov2022gpt4chan}.  Addressing the risk of abuse is vital to maximize the potential benefit of NLG technology, while minimizing harms --- a key principle of trustworthy AI \cite{doi/10.2759/346720}.

The overwhelming majority of contemporary state-of-the-art NLG models are neural language models (NLMs) based on the Transformer architecture \cite{Vaswani2017AttentionIA}.  Significant concerns surrounding the threats posed by generative Transformer models are nearly as old as the models themselves: the release of the 1.5B parameter GPT-2 architecture was delayed for nine months due to fears of abuse \cite{radford2019language}.  Access to GPT-3 remains only permitted via a carefully controlled API \cite{Brown2020LanguageMA}.  Such measures demonstrably manifest only in delays to open availability of models.  Only four months after the release of GPT-2, Grover --- a 1.5B parameter model based on the GPT-2 architecture --- was made publicly available \cite{zellers2019defending}.  The release of Grover not only foreshadowed the speed with which private models would be replicated, but also represented a limited threat model in itself: Grover was specifically designed to both produce and detect neural fake news.  Grover's primary author provided a reasoned justification for the model release, and called for an improved set of community norms for the release of potentially dangerous research prototypes \cite{Grover2019Release}.

Such norms have been slow to develop \cite{liang2022community-norms}, and wide scale democratization of access to increasingly large scale natural language generation models has continued. Open-source initiative EleutherAI has produced open-source generative Transformer models with large numbers of parameters, including the 6B parameter GPT-J \cite{gpt-j}, and 20B parameter GPT-NeoX \cite{black2022gptneox20b}. Even truly massive models are now available open-source --- the BigScience Large Open-science Open-access Multilingual Language Model (BLOOM) is an open-source multilingual model, and at 176B parameters, is larger than GPT-3 \cite{okBLOOMers}.  Yandex \cite{yalm}, Meta AI \cite{zhang2022opt}, and Huawei \cite{DBLP:journals/corr/abs-2104-12369} have all open-sourced models with over 100B parameters.

Real life examples of how generative Transformer language models may be  abused are beginning to emerge.  A controversy in the AI research community resulted from the publicized development of a GPT-J model trained on the 4chan politics message board \texttt{/pol/}.  This model was subsequently deployed to produce a large number of posts on the board from which its training data came, including posts containing objectionable content \cite{kurenkov2022gpt4chan}.  At it's peak, the model represented roughly 10\% of all activity on the board in a 24 hour period \cite{kilcher_2022}.  The response to the deployment of this model included a signed condemnation from 360 signatories across the AI community including scientific directors, CEOs, and professors \cite{liang_reich_2022}.  A similar project targeted a federal public comment website with GPT-2 text until the submitted comments made up half of all comments, demonstrating the extent of existing vulnerabilities \cite{weiss2019deepfake}.

Controversy around any individual publicized NLG model belies the more fundamental concern --- for years now, any person with access to adequate hardware and open-source training scripts could train or fine-tune large generative Transformers for any purpose they choose, be it pop song lyrics, mass disinformation, or toxic spam.  Malicious individuals in the process of training a generative language model need not draw attention to their models via public release, and currently face limited risk of discovery.  As NLG capabilities grow and access barriers evaporate, we are inevitably already quietly climbing the adoption curve for this technology to be widely abused by cybercriminals, disinformation agencies, scam artists, and other threat actors.


Access to these models is increasingly not limited to sophisticated threat actors who are able to fine-tune them.  User-friendly web interfaces, such as the one provided by ChatGPT \cite{openai_2022}, effectively eliminate any barrier to usage of powerful generative models.  Jasper, a tool marketed as an AI writing assistant, uses GPT-3 to write sections of content alongside a human's guidance \cite{jasper}.  This includes generating content for blogs and websites, which Jasper can efficiently produce in large volumes.  Another website offers an endless supply of GPT-3 authored cover letters \cite{open_cover_letter_2022}.  Tools such as Jasper allow those with little technical knowledge to seed the model with a prompt, specify keywords to include, and indicate a specific tone of voice.  Using publicly available open-source models, a nearly identical system could easily be created to generate endless streams of targeted disinformation, ready to be loaded into existing grey-market account automation tools for popular social media websites.

NLG models have the potential to have an immense and transformative positive impact on human society.  A staggering 1 in 3 internet users aged 16 to 64 have used an online translation tool in the last week, a figure representing over 1 billion people \cite{kemp_2022_cultural}.  Text summarization can create understandable summaries of complex legal text \cite{kanapala2019text} or medical records \cite{wang2021systematic}.  NLG models can give a voice to machine systems, changing the way that humans interact with them \cite{kim2019comparing}.  The same Transformer architecture used heavily for NLG can also be used for generating pictures from image descriptions \cite{ramesh2021zero}, producing functional code from a natural language summary \cite{chen2021evaluating}, and serves as the basis for the current vanguard of generalist agents \cite{gato}.  While future research in NLG will bring further wonders, alongside these opportunities is the corresponding certitude that the same technology will be used by bad actors to nefarious ends.  Predicting how abuses are likely to unfold, and understanding the best defenses against them, is essential for allowing humanity to reap the positive benefits of this technology while minimizing potential harms.  We must walk a cautious path through the age of the silicon wordsmith.

\subsection{Survey Overview}
\label{ssec:surveyover}

Since the release of GPT-2 \cite{radford2019language} and subsequent explosion of high-quality Transformer-based NLG models, there has been only one general survey on detection of machine generated text \cite{jawahar2020automatic}.  The scope of this previous survey is constrained to detection methods specifically targeting the several generative Transformer models that had been released at the time.  Prior to this, a systematic review of machine generated text predating the Transformer architecture covered approaches to detecting previous NLG approaches, such as Markov chains \cite{beresneva2016computer}.  Our survey differs from previous work in three major ways.

First, our survey of machine generated text detection is more comprehensive than previous work.  We consider literature on feature-based detection of machine-generated text that was omitted from prior review \cite{8282270, frohling2021feature, Lavoie2010AlgorithmicDO}.  Such approaches are a worthy inclusion as feature-based approaches still apply against contemporary NLG models \cite{frohling2021feature, crothers2022adversarial, kowalczyk2022detecting}, and may provide benefits such as improved robustness against adversarial attacks targeting neural networks \cite{crothers2022adversarial}, or enhanced explainability \cite{kowalczyk2022detecting}.  Additionally, as research on both NLG and detection has continued to rapidly advance in the years following the previous survey, we must now cover a wider range of generative models and defensive research.

Second, in addition to a comprehensive review of detection methods targeting contemporary models, this survey provides an in-depth analysis of the risks posed by NLG models via the process of \textit{threat modeling} (i.e., identifying potential adversaries, their capabilities and objectives) \cite{DBLP:journals/corr/abs-2107-01806}.  The result of our threat modeling process is a series of \textit{threat models} that describe scenarios where machine generated text may be abused, the likely methodology of attackers, and existing research related to each threat.  To date, there has yet to be any survey of machine generated text detection with a focus on the risks presented by machine generated text.  Consideration of threat models is vital to set the groundwork for trustworthy development of NLG technology, encourage early development of defensive measures, and minimize potential harms.

Third, guided by the EU Ethics Guidelines for Trustworthy AI \cite{doi/10.2759/346720} and research community efforts \cite{10.1145/3491209}, we present our survey with sociotechnical and human-centric considerations integrated throughout, focusing not only on NLG systems and machine text detection technologies, but on the humans who will be exposed to both text generation and detection systems in daily life.  The goal of trustworthy AI is to ensure that AI systems are developed in ways that are lawful, ethical, and robust both from a technical and social perspective.  Abuse of NLG models threatens all three of these areas, representing safety risks to those who may be targeted by NLG-enabled attacks, threats to the integrity of online social spaces, and challenges to the resilience of the technical and social systems that comprise modern society.  Machine text detection is part of protecting against abuse of NLG models, enhancing the robustness and safety of NLG development.  Critically, our survey also includes insight into ensuring defensive machine text detection systems themselves are transparent, fair, and accountable.  

To summarize, the major contributions of this work are as follows:

\begin{itemize}
    \item The most complete survey of machine generated text detection to date, including previously omitted feature-based work and findings from recent contemporary research.
    \item The first detailed review of the threat models enabled by machine generated text, at a critical juncture where NLG models and tools are rapidly improving and proliferating.
    \item A meaningful exploration of both topics through the lens of Trustworthy AI (TAI), considering the ethical and trust impacts of both threat models and detection systems.
\end{itemize}

The rest of this survey is organized as follows.  We provide definitions and a brief overview of existing methods for natural language generation in Section \ref{sec:nlg}.  In Section \ref{sec:threat} we explore threat models related to abuse of machine generated text, including impacts on trust.  We provide a comprehensive survey of literature related to detection of machine generated text in Section \ref{sec:detection}.  In Section \ref{sec:trends} we summarize open problems and ongoing trends to guide the direction of future work.  Finally, in Section \ref{sec:conclusion} we present our final conclusions.  While this work discusses machine generated text extensively, including models designed for generating scientific papers, no such models were utilized in authorship of this work.


\section{Machine Generated Text}
\label{sec:nlg}

Before reviewing threat models and detection methodologies for machine generated text, it is helpful to briefly provide a formal definition of machine generated text, and a condensed overview of natural language generation (NLG) models.  We recommend further reading of dedicated surveys on natural language generation for greater insight into the wide breadth of NLG models and applications \cite{10.1145/3554727, gatt2018survey, perera2017recent, santhanam2019survey, ijcai2021p612, reiterdale2002}.

\subsection{Definition and Scope}
In this survey, we use a broad definition of the term ``machine generated text" which we believe includes all relevant research in the field:

\medskip
\begin{quote}
    \centering
    \textit{``Machine generated text" is natural language text that is produced, modified, or extended by a machine.}
\end{quote}
\medskip

We focus our definition of machine generated text on \textit{natural language}  --- i.e., text written in human languages that are ``acquired naturally (in [an] operationally defined sense) in association with speech" \cite{lyons1991natural} --- and exclude \textit{non-natural language}  --- i.e., logical languages, programming languages, etc.  Exclusion of non-natural language aligns with other work in the field: the term ``text generation" is currently considered synonymous with ``natural language generation" \cite{ijcai2021p612, Zhang2009}.  We anticipate that ``text generation" may be repurposed in future research as an umbrella term that includes non-natural language text as well.  This would accommodate common considerations between NLG models and contemporary code generation models, such as Codex \cite{chen2021evaluating} and CodeGenX \cite{CodeGenX}.  As an example, attacks against StackOverflow or GitHub may include both NLG as well as vulnerable code generation.  Code generation models can also be used to complete programming assignments without triggering common plagiarism detection tools \cite{biderman2022fooling}. 

Our definition of machine generated text is intentionally broad, and covers a large number of possible use cases and associated threat models, which will be discussed in Section \ref{sec:threat}.  In the interests of managing a survey scope that already spans a wide range of literature and broad sociotechnical context, text generation by means of text adversarial attack will not be considered.  In the majority of cases, the production of new text is not the primary goal of a text adversarial attack, and text adversarial attacks and threat models are already covered by surveys in adversarial attack literature \cite{chakraborty2018adversarial, DBLP:journals/corr/abs-1902-07285, DBLP:journals/corr/abs-2005-14108}.  We will nevertheless discuss the role machine generated text plays in adversarial contexts in Section \ref{sec:threat}, as well as adversarial robustness of detection models in Section \ref{sec:trends}.

Note that this analysis focuses on threat models where a threat actor leverages machine generated text as part of an attack --- typically scenarios where the attacker is attempting to pass machine text as human, and where detection of machine generated text may be useful defensively.  We are not discussing attacks against NLG models themselves, unless they leverage NLG as part of the attack.  For example, a white-box training data extraction attack targeting the weights of a commercial speech-to-text model would not be included in our analysis, but using an NLG model to produce data for poisoning that model's training dataset would.

With this definition of machine generated text in mind, and with an understanding of the scope of research under consideration, we proceed to a brief overview of natural language generation.

\subsection{Natural Language Generation}

Using a computer to produce human-like text is well-established in the history of computing.  Turing's proposed ``imitation game" \cite{10.1093/mind/LIX.236.433} in 1950 considered the question of machine intelligence based on the ability of a machine to conduct human-like conversation over a text channel, for which the first widely-published method dates back to 1966 with the ELIZA chatbot \cite{10.1145/365153.365168}.  Given the large volume of NLG research over the past 55 years, we provide only a high-level taxonomy of major NLG tasks and approaches as groundwork for our analysis of threat models and detection methodologies, and leave detailed discussion to aforementioned dedicated surveys.

\subsubsection{Natural Language Generation Tasks}

Recall from \S \ref{ssec:risks} that there are a wide variety of applications for natural language generation.  Leveraging previous surveys \cite{ijcai2021p612, 10.1145/3554727, jin-etal-2022-deep}, we provide a summary of major tasks in the NLG domain, with examples of models that have been used for each task in Table \ref{tab:nlgapp}.  Note that many of the models listed are multi-purpose and can be trained for numerous NLG tasks.  In Table \ref{tab:nlgapp}, we provide a small selection of models that have been used for each task as representative examples.

\begin{table}[]
\footnotesize
\def\arraystretch{1.5}
\caption{Inputs, tasks, and example models for natural language generation}
\label{tab:nlgapp}
\begin{tabular}{@{}l|l|l@{}}
\toprule
Input                          & Task                          & Example models                \\ \midrule
None / Random noise            & Unconditional text generation & GPT-2 \cite{radford2019language}, GPT-3 \cite{Brown2020LanguageMA} (no prompt)  \\ \midrule
\multirow{6}{*}{Text sequence} & Conditional text generation   & GPT-2 \cite{radford2019language}, GPT-3 \cite{Brown2020LanguageMA} (with prompt), T5 \cite{2020t5}  \\ \cmidrule(l){2-3} 
                               & Machine translation           & FairSeq \cite{ott2019fairseq}, T5 \cite{2020t5} \\ \cmidrule(l){2-3} 
                               & Text style transfer           & Style dictionary \cite{sheikha2011generation}, GST \cite{sudhakar2019transforming} \\ \cmidrule(l){2-3} 
                               & Text summarization            & BART-RXF \cite{aghajanyan2021better}, Word and Phrase Freq. \cite{luhn1958automatic}  \\ \cmidrule(l){2-3} 
                               & Question answering            & FairSeq \cite{ott2019fairseq}, T5 \cite{2020t5}  \\ \cmidrule(l){2-3} 
                               & Dialogue system                 & DG-AIRL \cite{DBLP:journals/corr/abs-1812-03509}, DIALOGPT \cite{zhang2020dialogpt}, BlenderBot3 \cite{blenderbot3}, ChatGPT \cite{openai_2022}                     \\ \midrule
Discrete attributes            & Attribute-based generation    & MTA-LSTM \cite{feng2018topic}, PPLM \cite{Dathathri2020Plug}, CTRL \cite{Keskar2019CTRLAC}               \\ \midrule
Structured data                & Data-to-text generation       & DATATUNER \cite{Harkous2020}, Control prefixes (T5) \cite{clive2021control}                      \\ \midrule
\multirow{3}{*}{Multimedia}    & Image captioning              & GIT \cite{wang2022git}, ETA \cite{li2019entangled}                       \\ \cmidrule(l){2-3} 
                               & Video captioning           & MMS \cite{li-etal-2017-multi}, YouTube2Text \cite{6751448}                       \\ \cmidrule(l){2-3} 
                               & Speech recognition            & ARSG \cite{chorowski2015attention}, wav2vec-U \cite{baevski2021unsupervised} \\ \bottomrule
\end{tabular}
\end{table}

The summary in Table \ref{tab:nlgapp} is not exhaustive, and in reality, a mutually exclusive delineation between input types does not exist.  Combinations of different input types are possible.  As an example, CTRL takes both a discrete control code attribute and conditional text prompt in generation \cite{Keskar2019CTRLAC}.  Question-answering systems may be able to answer questions about images, such as Unified VLP \cite{zhou2020unified} and TAG \cite{https://doi.org/10.48550/arxiv.2208.01813}.  We consider a ``topic" as an attribute in this overview, and so include ``topic-to-text generation" under the broader umbrella of ``attribute-based generation", including work such as topic-to-essay generation \cite{feng2018topic}.

Given the strong generative capabilities of Transformer language models, and the corresponding increased risk of associated threat models, Transformer-based models rightly warrant particular emphasis in review.  However, as mentioned in Section \ref{ssec:surveyover}, consideration of the broader field of natural language generation and previous detection research is important as detection techniques that apply against pre-Transformer models have been shown to be useful in detection of modern generative models, and diverse approaches may offer increased adversarial robustness \cite{crothers2022adversarial} or better explainability \cite{kowalczyk2022detecting}.

\subsection{Natural Language Generation Approaches}

\begin{figure}[h]
\footnotesize
\centering
\caption{Taxonomy of major NLG approaches}
\vspace{1.5em}
\label{fig:nlg_approaches}
\begin{adjustbox}{center}
\begin{forest}
  for tree={
    parent anchor=south,
    child anchor=north,
    fit=band, 
  }
  [NLG Approaches
    [Neural, tier=neurality
      [VAE, tier=neural]
      [GAN, tier=neural]
      [RL, tier=neural]
      [IRL, tier=neural]
      [RNN, tier=neural
        [LSTM, tier=rnn]
        [GRU, tier=rnn]
      ]
      [Transformer, tier=neural
        [GPT-*, tier=transformer]
        [Grover, tier=transformer]
        [BART, tier=transformer]
        [CTRL, tier=transformer]
      ]
    ]
    [Non-Neural, tier=neurality
      [Rule-based, tier=nonneural]
      [Statistical + Rule-based, tier=nonneural
        [K-means, tier=statistical]
        [HMM, tier=statistical]
        [SVM, tier=statistical]
      ]
      [MDP RL]
    ]
  ]
\end{forest}
\end{adjustbox}
\end{figure}
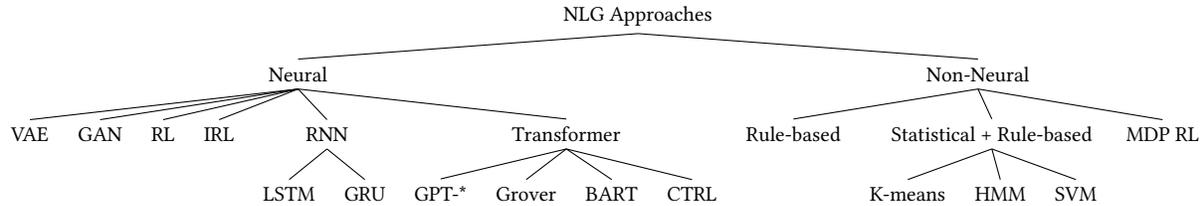

There are a wide range of model architectures and algorithmic approaches to natural language generation.  We categorize these approaches broadly into neural and non-neural methods, and then further break them down into more specific categories.  A diagram of our simplified breakdown can be found in Figure \ref{fig:nlg_approaches}.  As previously mentioned, NLG encompasses a large variety of tasks and research areas, with this brief section serving as context for understanding machine generated text threat models and detection methods.

\subsubsection{Non-Neural Models}

Predating the popularization of neural approaches in the NLG domain, a range of systems were used to accomplish NLG tasks.  These early approaches can broadly be summarized as ``rule-based", though there existed variety in terms of processes, pipelines, and targets tasks.   A review of rule-based systems can be found in Reiter and Dale's book on the subject \cite{reiterdale2002}.

An alternative approach to purely rule-based approaches is to use an existing natural language corpus to generate rules for components of an NLG system, such as content selection \cite{duboue-mckeown-2003-statistical, langkilde1998generation} or template generation \cite{kondadadi2013statistical}.  These statistical approaches are meant to be more adaptable to different domains than strictly rule-based systems.  While many different statistical models have been integrated with NLG systems in various ways, Hidden Markov Models (HMM) \cite{baum1966statistical} feature prominently in past work.  More recent non-neural research has used reinforcement learning \cite{janarthanam2009learning} and hierarchical reinforcement learning \cite{dethlefs-cuayahuitl-2010-hierarchical} of Markov Decision Process (MDP) agents to learn optimal text generation policies.

\subsubsection{Non-Transformer Neural Methods}

Natural language generation using neural networks was demonstrated to be highly effective using recurrent neural networks (RNN) \cite{kalchbrenner2013recurrent, mikolov2012statistical, Berglund2015BidirectionalRN}, including long short-term memory (LSTM) architectures \cite{merity2017regularizing} and gated recurrent units (GRUs) \cite{Pawade2018StoryS}.  However, RNN and LSTM architectures had to contend with the vanishing gradient problem, to which the multi-head attention mechanism of the Transformer architecture is more resilient \cite{Topal2021ExploringTI}.  Generative adversarial networks (GANs) \cite{goodfellow2020generative} --- commonly used to generate continuous data (such as images) --- can also be adapted to a discrete context for natural language generation \cite{Lin2017AdversarialRF, Yu2017SeqGANSG}.

Deep reinforcement learning (RL) has been used with neural networks to learn policy gradient methods that reward text characteristics associated with high-quality text generation \cite{Li2016DeepRL}.  A related area of work is the usage of inverse reinforcement learning (IRL), which has included work that aims to address reward sparsity and mode collapse problems in GAN-based text generation by learning an optimal reward function and generation policy \cite{DBLP:journals/corr/abs-1812-03509, Shi2018TowardDT}.

\subsubsection{Transformer}

\label{ssec:transformer}

The multi-head attention architecture of Transformer language models \cite{Vaswani2017AttentionIA} currently represents the state-of-the-art in natural language generation across natural language tasks.  Among Transformer models, the unidirectional GPT-2 \cite{radford2019language} and GPT-3 \cite{Brown2020LanguageMA} models are the most studied in the field of machine generated text detection due to their groundbreaking performance on unconditional and conditional text generation --- though like many other Transformer models, these architectures can be used for other NLG tasks as well.

In addition to GPT-2 and GPT-3, also notable are related autoregressive language models using similar architectures, with variations in sampling procedures or training datasets.  Such models include Grover \cite{zellers2019defending} (a GPT-2 style model trained on a news dataset and using nucleus sampling instead of top-$k$ sampling), GPT-J \cite{gpt-j} (a 6-billion parameter autoregressive language model trained on The Pile \cite{DBLP:journals/corr/abs-2101-00027}), and GPT-NeoX-20B \cite{black2022gptneox20b} (a 20-billion parameter model similar to GPT-3, also trained on The Pile \cite{DBLP:journals/corr/abs-2101-00027}).

Unidirectional Transformer language models generate text by performing self-supervised distribution estimation to predict the next token based on previous tokens.  To do this, the model is trained on an existing set of variable-length example texts $(x_1, x_2, ..., x_n)$ each composed of symbols $(s_1, s_2, ..., s_m)$.  These symbols may be characters, or multi-character tokens obtained through a tokenization process.

The probability of a given text can then be expressed as the conditional probability of the final token, given each previous token.  That is:

\begin{equation}
p(x) = \prod_{i=1}^{m} p(s_m|s_1,...,s_{m-1})
\end{equation}

The self-attention mechanism in the Transformer architecture makes it possible to train neural network architectures that are effectively able to estimate such probabilities, given a suitable pre-training task.  In unidirectional models such as those in the GPT lineage, a common training task is prediction of the next token in sequence.  To generate text, such models can then receive a continue an input sequence by sampling from the probability distribution of all possible next tokens based on previous tokens.  An important parameter in this sampling process is ``temperature" $T \in (0, \infty)$, which can be raised above 1 to increase the likelihood of selecting a less-probable next token --- improving diversity at the potential cost of choosing an unusual token --- or lowered below 1 to bias sampling towards more common tokens.

There are three common decoding strategies used for sampling token probabilities from contemporary unidirectional generative Transformer models \cite{DBLP:journals/corr/abs-1904-09751}:

\begin{enumerate}
    \item No truncation $\rightarrow$ Sample from the entire probability distribution.  At $T=1$, this is called ``pure sampling".
    \item Top-$k$ truncation $\rightarrow$ Sample from the $k$ most probable tokens.
    \item Nucleus sampling (also known as top-$p$ truncation) $\rightarrow$ Sample from tokens in the top-$p$ portion of the probability mass, rather than a fixed number of tokens $k$.
\end{enumerate}

Alternative methods of sampling are an active area of research in improving text generation.  Such methods include `typical sampling', in which tokens are selected based on expected information gain, rather than strictly probability of occurrence \cite{10.1162/tacl_a_00536}.

While unidirectional generative models are key fixtures of machine generated text detection research, other Transformer architectures can be used for NLG tasks as well.  The architecture of BART \cite{lewis-etal-2020-bart} includes a bidirectional encoder (similar to BERT \cite{DBLP:journals/corr/abs-1810-04805}), but maintains a left-to-right decoder for sequential text generation.  Other Transformer architectures such as MASS \cite{song2019mass}, T5 \cite{2020t5}, and ULMFiT \cite{howard2018universal} can also be used for NLG tasks.

An important area of ongoing research centers around shaping the output produced by Transformer models.  This can include prompt engineering --- carefully crafting the conditional text input for a language model to continue \cite{Brown2020LanguageMA} --- or by providing additional discrete attributes that can be used to influence the generation of the network, such as control code, topic, or sentiment as in CTRL \cite{Keskar2019CTRLAC}, PPLM \cite{Dathathri2020Plug}, or GeDi \cite{krause2020gedi}.  Greater control over model output increases the risks posed by threat models \cite{Brown2020LanguageMA}.  As an example, when generating social media posts as part of an NLG-augmented online influence campaign, an attacker would benefit from being able to ensure that generated comments both 1) mention a targeted political opponent, and 2) demonstrate negative entity sentiment  towards the opponent.  We will cover such potential abuses and others in more detail in the next section, which concerns threat models associated with machine generated text.

\section{Threat Models}
\label{sec:threat}


Machine generated text enables a diverse array of attacks.  These attacks may be performed by threat actors with specific objectives, such as to compromise a computer system, exploit a target individual for financial gain, or enable large-scale harassment of specific communities.  The EU ethics guidelines for trustworthy AI emphasize that unintended or dual-use applications of AI systems should be taken into account, and that steps should be taken to prevent and mitigate abuse of AI systems to cause harm \cite{doi/10.2759/346720}.  As such, trustworthy AI in the context of NLG necessitates understanding the areas where such models may be abused, and how these abuses may be prevented (either with detection technologies, moderation mechanisms, government legislation, or platform policies).  When discussing attacks, we discuss not only the direct impact on targets, but also the broader impacts of both attacks and mitigation measures on trust.

To understand the risks that motivate research on detection of machine generated text, we draw from existing literature to present a series of threat models incorporating natural language generation.  Threat modeling reflects the process of \textit{thinking like an attacker}, identifying vulnerabilities to systems by identifying potential attackers, their capabilities, and objectives.  The goal of threat modeling is to improve the security of systems by considering the greatest threats to systems and their users.  Many methods of threat modeling have been developed over the years, including producing system diagrams, stepping through itemized vulnerability checklists, and performing open-ended brainstorming \cite{ncsc_2022, bromander2016semantic, kohnfelder1999threats, ucedavelez2015risk}.  In late 2020, a diverse set of experts formed a threat modeling working group to produce a high-level set of guidelines related to effective threat modeling approaches \cite{threatmodel_2020} --- we leverage these guidelines in the open-ended attack-centric modeling approach in this section.

\subsection{Threat Modeling Fundamentals}

As we anticipate an audience with varying exposure to cybersecurity topics, before we present threat models related to machine generated text, it is helpful to first provide an overview of threat modeling, and characterize the approach taken in this section.

A basic example of a common threat model is ``a thief who wants to steal your money" \cite{shostack2014threat}.  We can add detail to this threat model by considering more specific capabilities and objectives that such an attacker might have.  For example, we may consider ``a thief with lockpicks who wants to steal your TV", or ``a thief who found your banking password in a database dump and wants to transfer money out of your account".  With these threat models in mind, we can then propose mitigation strategies, such as ``install locks that are resistant to lockpicking", or ``use multi-factor authentication for online banking".  Finally, we evaluate whether our mitigation approach is sufficient to address the threat, and consider what other threat models we might need to consider.  Threat modeling is inherently an iterative process \cite{shostack2014threat, threatmodel_2020}.

\textit{Shostack's Four Question Frame for Threat Modeling} \cite{shostack2014threat, 4qgithub} presents best a plain language foundation for threat modeling by posing four simple questions:

\begin{enumerate}
\item \textit{What are we working on?} $\rightarrow$ Identify the system under attack.
\item \textit{What can go wrong?} $\rightarrow$ Determine potential attackers, their capabilities, and objectives.
\item \textit{What are we going to do about it?} $\rightarrow$ Devise a mitigation strategy.
\item \textit{Did we do a good job?} $\rightarrow$ Review whether the analysis is accurate and complete.
\end{enumerate}

In these terms, we summarize our threat modeling approach in this section as follows:

\begin{enumerate}
\item \textit{Identify the system under attack:} We provide a broad attack-centric analysis of machine generated text on society, rather than a system-centric analysis focusing on vulnerabilities to a specific IT system.  As such, we identify several discrete technological systems, within the broader societal supersystem.

\item \textit{Determine potential attackers, their capabilities, and objectives:} We consider threat actors of varying sophistication and motives, but with a common modus operandi --- in all cases, our attacker is an individual or organization exploiting an NLG model.  We characterize the attacker when explaining each attack.

\item \textit{Devise a mitigation strategy:} After identifying a threat model, we propose mitigation measures to improve security and reduce risk.  Detection of computer-generated text is often part of the presented mitigation approaches, but policy changes and human moderation systems can also have a significant impact.

\item \textit{Review whether the analysis is accurate and complete:} We have given careful thought to the presented threat models, which are formed from perspectives gained across industry, academia, and government. However, as threat modeling is an iterative process that benefits from diverse perspectives \cite{threatmodel_2020}, we greatly encourage further analysis of potential attacks and mitigation measures in future research.
\end{enumerate}



\begin{figure}[h]
  \centering
  \includegraphics[width=\linewidth]{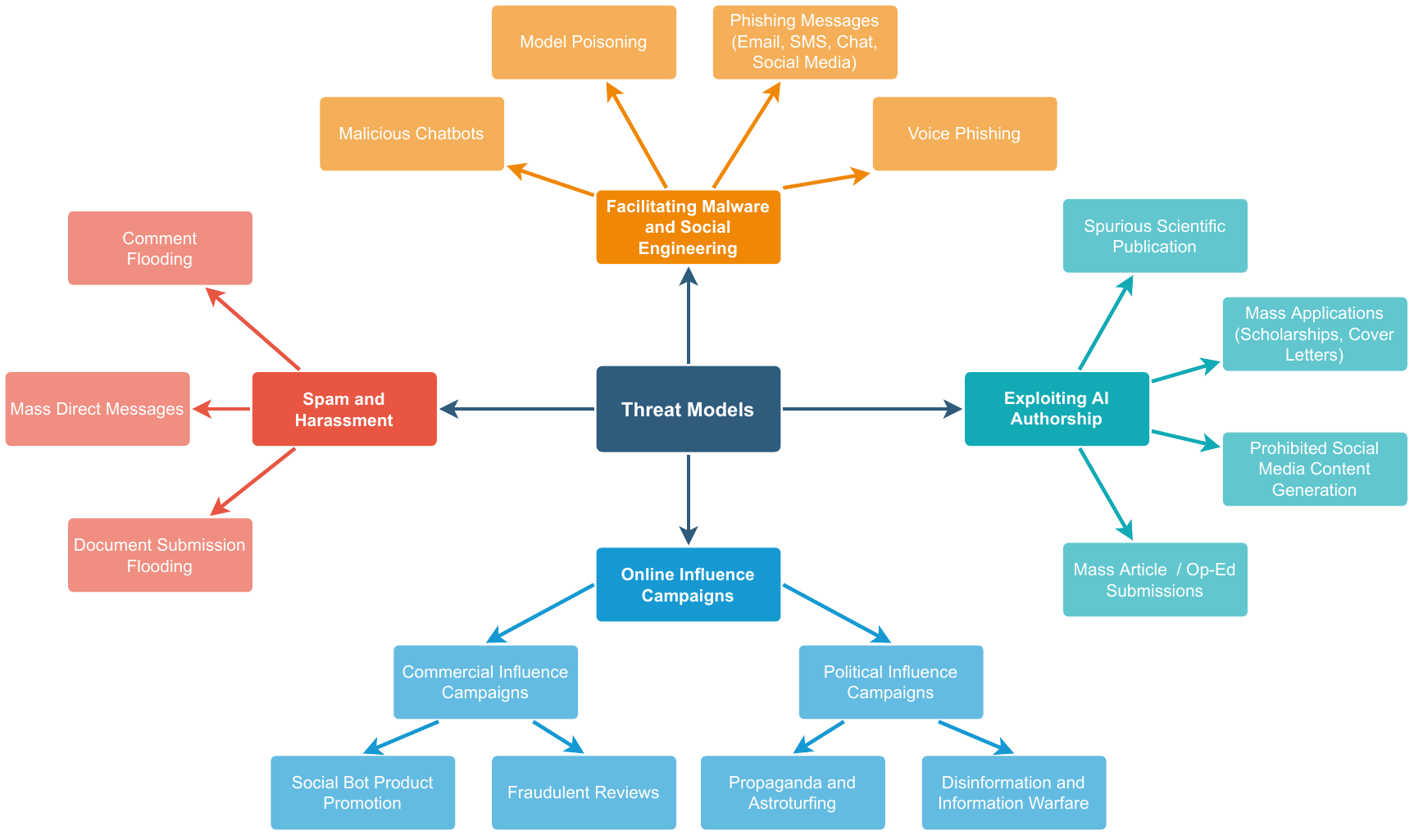}
  \caption{Broad taxonomy of threat models enabled by machine generated text}
  \label{fig:threat_models}
  \Description{A diagram depicting threat models mentioned in this research, broken down into broad categories.}
\end{figure}

The remainder of this section comprises our threat model analysis, grouped according to a breakdown of attacks into four major categories, followed by a concluding discussion.  Within each category we discuss threat models associated with that category of attack, identifying systems at risk, and describing possible threat actors, their objectives, and capabilities.  For each attack, we propose mitigations, and then discuss the trust impacts of both the attack and --- crucially --- of the proposed mitigations as well.  A taxonomy of the broad categories of attacks using NLG models we discuss can be found in Figure \ref{fig:threat_models}.



While a completely exhaustive list of all possible future malicious applications of NLG models is not possible, the threats outlined here span a wide range of tangible dangers at this point in time, representing valuable areas of future investigation for preemptive ethical defensive research.  As previously mentioned, threat modeling is iterative, and it is hoped that these threat models should serve as the foundation for future work in improving security against machine generated text.


\subsection{Facilitating Malware and Social Engineering}

\subsubsection{Phishing and Scamming}


Phishing attacks center on socially engineering a target individual to perform a desired action. This might be to convince the target to open an unsafe document that contains an exploit, cause the target to navigate to a fake banking webpage, encourage them to share sensitive information that can be used for identity theft, among many other documented methods  \cite{CHIEW20181, 10.3389/fcomp.2021.563060}.  Phishing attacks can target numerous channels, including email, phone, SMS, or chat applications.  

%



Automated messaging approaches in the early stages of phishing campaigns are common \cite{10.3389/fcomp.2021.563060}. Machine generated text can be a useful tool to an attacker attempting to scale or better target phishing or scam campaigns.  Rather than provide the same message to all targets, NLG can be used to generate target-specific text.  Research has demonstrated the effectiveness of NLG both for scaling of email masquerade attacks \cite{10.1145/3052973.3053037}, and for community-targeted phishing \cite{10.1007/978-3-030-14687-0_8}.  Carefully targeting a phishing attack (commonly referred to as ``spear phishing"), greatly increases the likelihood of a specific target falling for the attack \cite{doi:10.1080/10919392.2019.1552745}.  In the cases of chat messages, NLG models that serve as dialogue agents may be exploited to exchange messages with the target under a pretext before exploiting them \cite{zhang2020dialogpt}.  

Mitigation of NLG-enabled phishing attacks will be similar to established work on existing phishing attacks, including both automated detection systems, user reporting, and awareness campaigns \cite{abu2007comparison}.  NLG may present an increased challenge for existing detection systems in that generated messages may have unique or highly-varied content --- though attackers may be forced to include specific ``payload" content for an attack to be effective (e.g., a phishing email may include a unique shortlink to the same fraudulent website, a malicious chatbot may need to socially engineer responses to the same security questions).  As text content becomes more varied due to more powerful NLG models, detection of payload content may represent a stable detection feature. Algorithms for detection of machine generated text are also likely to be added to existing automated detection approaches.


\subsubsection{Social Worms}


NLG models may be particularly useful for worms that spread through social media or email contact networks.  When an individual has their account compromised by an exploit, that account may be used to send malicious messages that propagate the exploit to other users.  By using previous messages or emails between individuals as context to an NLG model, it may be possible to automatically produce messages that include personal details, mimic a loved ones writing style, or carry on a short conversation before delivering a malicious file or link.  Given that NLG models are often quite large, the NLG component of this may need to be run on a separate command-and-control server and queried from behind a proxy, rather than bundled with the exploit code itself (unless the pretext of the conversation can be used to convince the target to download a file).

Mitigation of such attacks could involve platforms adopting formal policies that users do not use machine generated text in their communications, except under carefully controlled circumstances.  Detection models could then be leveraged against user communications.  While this may be acceptable for public posts, there exists privacy risks when considering private messages.  Detection models could perhaps be executed as part of the message viewing application on the receiving device to protect end-to-end privacy of messages.  If a user receives several messages that score highly for machine generated text detection, a warning may be raised.  This approach is not without risks, as the privacy of direct messages must be protected, and any real or perceived erosion of privacy will undermine public trust.  Other security measures to protect accounts from unauthorized logins, such as multi-factor authentication, should continue to be used to protect against account compromise more broadly.

\subsubsection{Model Poisoning}

Cybercriminals may have an interest in poisoning the training datasets of machine learning models.  This may be to support other attacks (e.g., poisoning a training dataset for a malware detection algorithm or email spam filter) or poisoning a given model may be the primary goal (e.g., poisoning the dataset of an algorithmic trading model so that the attacker can later trigger trades that financially benefit the attacker).  If a threat actor identifies that they can access the training data of any such algorithm, they may use NLG models to produce many training examples containing a particular malicious signature they wish to conceal.  Poisoning attacks against neural code-completion algorithms have been performed by generating samples including a given vulnerability \cite{schuster2021you}, and GPT-2 has been used in research to produce fake cyber threat intelligence reports for poisoning cyber-defense systems \cite{9534192}.

Mitigation of dataset poisoning varies based on the sensitivity of the model and nature of the training dataset.  The first line of defense is basic IT security best practices that prevent unauthorized modifications to training datasets.  However, in some situations models are trained on publicly available data, and therefore it is not possible to prevent access to training data.  In these cases, data might be screened prior to inclusion in the training dataset.  This screening can include classifiers --- potentially including the machine generated text detection approaches discussed in Section \ref{sec:detection} --- or by other analysis methods, such as cluster-based methods to detect poisoning in training datasets \cite{baracaldo2017mitigating}.  Finally, for sensitive models, it may also be appropriate to leverage data versioning techniques and audit logging to capture changes to the data potentially made by a malicious insider.


\subsubsection{Impacts of Attacks and Mitigation on Trust}


The usage of NLG models to produce compelling, target-specific messages as part of large scale phishing attacks and social worms is likely to further reduce trust in text communications, particularly those received from contacts not personally known.  Individuals may become even more suspicious of unsolicited messages --- even seemingly innocuous ones.  As even a seemingly good-natured greeting may be just the first message from a malicious dialogue agent, individuals may decide it is safer to not reply to such messages, further reducing trust and social interaction with new individuals in online communities.

NLG-based poisoning attacks against machine learning models will likely have the greatest trust impact on machine learning practitioners, who may be required to carefully scrutinize open-source training data for poisoned samples.  Where mitigation of poisoning attacks involves limiting access to training datasets behind auditing and approval processes, such procedures may cause developers to feel distrusted, and undermine the relationship between these individuals and the organizations they work with.  While the trust impact of NLG-based data poisoning attacks may be relatively minor among the general population, a high-profile attack (e.g., a poisoning attack against a medical diagnosis model) may cause individuals to lose trust in machine learning systems more broadly, based on concerns that such models are not be safe from malicious tampering.

\subsection{Online Influence Campaigns}

An area of particular concern for abuse of machine generated text is facilitation of online influence campaigns.  The objectives of threat actors in this area may either be political in nature (e.g., disinformation, propaganda, election interference) or commercial in nature (e.g., product promotion, smearing competitors, fake reviews).  In either case, the goal is to promote a particular idea or prompt a particular action among the target audience.

Either type of campaign may both leverage or facilitate other threat models, such as spam, harassment, mass submission of agenda-driven content, phishing, or malware.  The distinction between commercial and political influence campaigns is useful for better understanding threat actors and threatened systems in more detail, as well as categorizing existing research.  

\subsubsection{Political Influence Campaigns}

Machine generated text as part of political influence campaigns has been analyzed in previous work \cite{shu2020mining, stiff2022detecting, zellers2019defending}.  Papers related to the threat of generative language models on online influence operations may use terminology such as terminology ``fake news" \cite{zellers2019defending} or ``disinformation" \cite{brief2021ai, stiff2022detecting}, or ``domestic and foreign influence operations" \cite{brief2021ai}.

The threat actor in a political influence campaign represents an entity who wishes to influence beliefs or prompt action among a target group. These threat actors might include, as examples:

\begin{itemize}
\item A political party hiring a group to post unflattering comments online about their political adversaries
\item A nation-state disseminating fraudulent news reports to mask human rights abuses
\item A nation at war attempting to incite the citizenry of an opposing nation to overthrow the government
\end{itemize}

Datasets from past political influence campaigns have been released by Facebook \cite{schrage_ginsberg_2018}, Reddit \cite{redditsuspicious}, and Twitter \cite{twitterei}, including operations attributed to 22 different countries.  Threat actors in this space can be expected to already have the capability of running online political influence campaigns using human employees or contractors, and be familiar with social media automation tools that facilitate registration and management of fraudulent social media accounts.  Large models with strong few-shot performance can simply be provided examples of the desired messaging to produce further propaganda.  We demonstrate this in Table \ref{tab:gpt3fewshot}, using Russian information operations executed on Twitter during the Syrian civil war to generate additional `on-message' tweets that promote Syrian Arab Army (SAA) and Russian forces, while discrediting American involvement.

\begin{table*}[!ht]
\centering
\small
\caption{Few-shot generation of information warfare based on Syrian civil war influence operations}
\label{tab:gpt3fewshot}
\begin{tabular}{@{}ll@{}}
\toprule
\begin{tabular}[c]{@{}l@{}}GPT-3 Prompt:\\Human tweets\\from Russian\\foreign military\\intelligence\\(GRU) \cite{twitterei}\\\end{tabular} & \begin{tabular}[c]{@{}l@{}}
RT @TheArabSource: Confirmed: \#US retrains \#ISIS and \#ISIS-like jihadists in east \#Syria and \\re-brands them 'democratic forces' (VIDEO) - E...\\\hline
RT @Tasnimnews\_EN: \#Iraq Army Discovers \#US Drone in \#Daesh Arms Depot near Fallujah\\\url{https://t.co/zYRgEHmlr0} \url{https://t.co/H762UxwlQx} \\\hline
...\\\hline
ISIS-affiliated band group Wilayat al-Sham recruits new \#terrorists in US-controlled \#Rukban camp.
\end{tabular} \\\midrule
\begin{tabular}[c]{@{}l@{}}GPT-3 Response\\\\\\\\\\\\\end{tabular}  & \begin{tabular}[c]{@{}l@{}}@ALFaham\_ \#SAA seized a big amount of \#ISIS \#ISIL weapons from the convoy in east \\\#Homs countryside today. https://t.co/k3kzPoLlsd
\#Syria \#Damascus \#Jobar \#ISIS\\\hline
\#Washington is willing to perform the functions of \#MiddleEastern policeman and play the role \\of accessory to \#Turkish\#aggression against \#Syria.
\url{https://t.co/HVGXuvRZ}\\\hline
...\\\hline
\#Russia's \#Air\_Force has stepped up airstrikes against camps for the \#Islamic\_State in the \#Terek\_Val
\end{tabular} \\\bottomrule
\end{tabular}
\end{table*}





As consumption of text content is common online, there are many avenues where machine generated text might be utilized by a threat actor to improve scaling and targeting of influence operations.  Social media, due to the large volume of engaged users, is likely to continue be a valuable and vulnerable target for such campaigns \cite{redditsuspicious, twitterei, schrage_ginsberg_2018}.  Research on detection of machine generated text for political influence campaigns that focuses on `fake news' only covers a very small subset of critical threat models.  Fake news detection research often imagines an adversary using an NLG model to produce news-like disinformation at scale \cite{zellers2019defending}.  Producing massive volumes of news-like content may be a less desirable machine generated text disinformation scaling approach than social messages for several reasons:

\begin{itemize}
\item Research has demonstrated that individuals are more likely to share an article than read it \cite{gabielkov:hal-01281190}, and that a majority of individuals make up their minds on news topics by only reading headlines \cite{author_2014}.
\item Scaling by number of articles does not multiply effectiveness --- a single news article or handful of news articles can be widely disseminated, reducing the need to generate large numbers of articles each day. 
\item Scaling by number of articles requires either manipulating existing platforms to host them (i.e., layering and information laundering \cite{meleshevich_schafer_2018}), or procuring domain names and hosting infrastructure, representing additional cost and effort.
\item Human involvement in fake news article authorship allows disinformation threat actors to better tailor messaging, reduce detection, and more carefully walk the line of promoting manipulative information without triggering moderation from social media websites 
\end{itemize}

Research on GPT-3 has shown that it and similar models can easily be used in a few-shot setting to generate conspiratorial content to promote radicalization \cite{DBLP:journals/corr/abs-2009-06807}.  However, in light of the above considerations, NLG usage by disinformation agencies for newswriting may focus on leveraging AI writing assistants such as Jasper to save time and minimize cost \cite{jasper}, and translation models to more effectively cross language barriers.  NLG models may be used for disinformation by operating social bots that distribute links to disinformation articles, promote discussion around incendiary headlines, and produce large numbers of comments that give the false impression of a public consensus.  Targeted users need not even read the shared articles --- the artificial amplification of a headline and overwhelming ``grassroots" narrative guided by machine generated comments is likely sufficient to influence public opinion \cite{gabielkov:hal-01281190, author_2014}.



Regarding mitigation, past research has identified that the average user is overly trusting of profiles with AI-generated photos and GPT-2 text, accepting connection requests from deepfake profiles on LinkedIn 79\%–85\% of the time \cite{279946}.  As such, it is unlikely that user reports will serve as an adequate first line of defense.  Instead, a combination of automated detection models (including machine generated text detection) and platform moderation efforts should be used to detect political influence campaigns. Among these should be measures to protect against social media abuse more broadly, including detection of account automation, and scrutiny of coordinated inauthentic activity for content amplification.  Investigations by disinformation researchers, such as those carried out on Twitter, are likely to remain relevant \cite{twitterei}.

\subsubsection{Commercial Influence Campaigns}


In commercial influence campaigns, the goal is to influence individuals in a manner that commercially benefits the threat actor.  Examples of such campaigns include publishing fraudulent reviews, artificially boosting a website's search engine page ranking, spamming online communities with advertisements for a product, or attempting to inorganically cause promotional content to trend on social media.  As with previous categories, there may be overlap between different attacker approaches.

A threat model of particular interest is the usage of machine generated text to generate fraudulent reviews that either promote one's own product/service, or target a competitor \cite{stiff2022detecting, 10.1007/978-3-030-44041-1_114, kowalczyk2022detecting}.  Work has been published that demonstrates sentiment-preserving fake reviews, which might be used for such a purpose \cite{10.1007/978-3-030-44041-1_114}.  Fake reviews can be abused on marketplace websites themselves, or by targeting potential customers on social media platforms.  Threat actors may operate such campaigns themselves, or may avail themselves of the thriving online market for fake reviews \cite{he2022market}.  Organizations selling fake reviews may become early adopters of open-source NLG models to provide unique and specific reviews at lower cost.


Mitigation of NLG models used for fake reviews on online marketplaces might involve running machine generated text detection on the text of reviews, in addition to other detection systems currently used to combat this problem.  Advanced NLG models should not affect context-based detection methods (e.g., identifying patterns in reviewer usernames, similar account creation times, unusual purchase behaviour, etc.).  It may be more difficult to detect commercial influence campaigns if attackers post content outside marketplace websites.  For example, social media websites (e.g., Facebook, Instagram, Reddit, YouTube comments), map platforms (e.g., Google Maps), or dedicated review sites (e.g., Yelp) may all be locations where false reviews may be posted.






\subsubsection{Impacts of Attacks and Mitigation on Trust}

In addition to the risks posed by machine generated text for online influence campaigns, the existence of NLG threat models causes additional damage to trust online.  The perception that any given user on social media may be a bot, can cause users of social media to dismiss others (particularly individuals whom they don't agree with) as ``bots", rather than acknowledge that other real people may hold different viewpoints.  The net effect of this is reduced trust in the authenticity of online interactions.  


Mitigation of NLG-enabled influence operations via automated detection of machine generated text also itself carries potential negative impacts.    Automated detection creates the possibility of mass-suppression of speech online.  Previous work has found that text written by non-native English speakers that included political topics was of high risk of being erroneously detected by a Transformer trained on previous political influence campaigns \cite{crothers2019towards}.  As methods based on RoBERTa (also a Transformer) are currently the state of the art for detection of machine generated text \cite{Liu2019RoBERTaAR, solaiman2019release}, classifiers for machine generated text detection leveraged to combat online influence campaigns must be carefully trained and ethically evaluated to minimize the risk of similar incidences of mass discrimination.  Continued public reporting of influence campaign datasets, such as the regular releases by Twitter for review by researchers \cite{twitterei}, would be beneficial to protecting trust in social media moderation.


Language background considerations evoke another problem: there are legitimate reasons why a user may rely on machine generated text.  A person writing in their non-native language may leverage an online translation model to assist them.  While such text may be considered machine generated text, this text is not \textit{inauthentic} --- it nevertheless represents genuine self-expression. Much of the world relies on translation tools to better participate in online discourse; recall that 1 in 3 internet users aged 16 to 64 have used an online translation tool in the last week \cite{kemp_2022_cultural}.  Relying on machine generated text detection alone is therefore likely to produce a solution that is discriminatory, unreliable, and greatly damages trust in social media platforms.  Machine generated text detection should then be used among multiple features, such as account creation times, activity patterns, registered phone numbers, and IP addresses, to determine whether activity is linked together as part of an online influence operation.



\subsection{Exploiting AI Authorship}

\subsubsection{Academic Fraud}
\label{sssec:academicfraud}

Use of algorithms to generate scientific papers has been well-established since SCIgen was created in 2005 to produce nonsensical papers that nevertheless sometimes passed peer review \cite{hargrave_2005}.  These papers continue to emerge in respected publications, many years later, despite the comparative simplicity of the context-free grammar generation method \cite{labbe2013duplicate, cabanac2021prevalence}.  Generation of artificial scientific papers uses up valuable reviewing resources, lowers publication quality standards by producing misleading or nonsensical publications, and challenges trust in the scientific review process itself.  In education, NLG models may be used by students to cheat on language learning assignments via machine translation \cite{song2019mass}, or easily produce essays on a given topic \cite{feng2018topic, dehouche2021plagiarism} --- both instances where institutions may be tempted to perform detection of machine generated text to improve academic integrity and encourage students to learn course material.  Widespread access to convenient NLG interfaces online, such as that provided by ChatGPT \cite{openai_2022}, allow any student with an internet connection to leverage such models, even when doing so undermines the learning objectives of an assignment (i.e., cheating).

Threat actors submitting AI-generated papers are typically either 1) academics attempting to inflate publication statistics, particularly when meeting a quota in order to maintain their position \cite{cabanac2021prevalence}; or 2) well-meaning researchers probing the publication standards of a potentially disreputable conference \cite{10.1145/1255175.1255220}.  Capabilities of threat actors include usage of well-established tools such as SCIgen, or usage of more recent Transformer-based approaches that are promoted as ``scientific writing assistants" which can nevertheless be easily exploited to generate long articles of little substance \cite{10.1007/978-3-030-62327-2_27}.  Mitigation measures should include flagging likely machine-authored publications using published approaches for detection of SCIgen articles \cite{labbe2013duplicate, cabanac2021prevalence}, as well as new detection approaches based on detection of Transformer generated text \cite{rodriguez-etal-2022-cross}.  Human reviewers can more carefully review flagged articles to determine whether the article contains credible research, irrespective of the detection result.

The acceptability of machine text within scientific writing is an active area of discussion in academic disciplines.  The Association for Computational Linguistics has released a set of guidelines on the usage of AI writing assistance \cite{graber_okazaki_rogers_2023}. If the results published by a researcher are true and accurate, limited usage of a carefully guided NLG model may be considered acceptable by some publications.  Research has been emerging that aims to differentiate between acceptable and unacceptable usage of NLG models in scientific writing \cite{https://doi.org/10.48550/arxiv.2209.03742}, which should be part of a broader ongoing social conversation on norms surrounding AI usage and disclosure. 

\subsubsection{Applications and Cover Letters}

Contemporary NLG models can be used to generate large numbers of cover letters or essays for applying for scholarships or to employment opportunities.  Commercial websites already exist for producing cover letters using GPT-3 \cite{open_cover_letter_2022}.  While the overall usefulness of human-written cover letters has been debated in business media \cite{lufkin_2021}, they are ostensibly meant to be an earnest reflection of a candidate.  Usage of AI models to generate a cover letter or essay submission is therefore likely to be considered exploitative by organizations who review such submissions.  The threat actors in this case may be individuals (perhaps understandably) looking to save time and improve their employment opportunities by bypassing a cumbersome application process, or a malicious attacker looking to flood a target company with fraudulent submissions (a threat actor which we will discuss further in ``Spam and Harassment").

Detection of machine generated text may be able to identify artificial cover letters or essays given they are of sufficient length (the odds of successful detection improve with sequence length \cite{Ippolito2020AutomaticDO, zellers2019defending, radford2019language}).  However, caution should be taken with this approach, as use of AI writing tools is not necessarily exploitative.  Again, individuals writing in a second or third language may rely on translation models or NLG writing assistants to help them write cover letters or scholarship applications.  It may be difficult to differentiate those who mean to exploit such systems (e.g., thoughtlessly spam submissions to as many avenues as possible), and those who are relying on AI writing tools to better express themselves.  As such, a better mitigation approach may be to develop alternative approaches to evaluating candidates, such as placing more emphasis face-to-face discussions with prospective job candidates or award recipients.

\subsubsection{Content Generation}

A growing threat model for social media platforms is the possibility that a large number of users may begin using generative AI models (including NLG models) to produce social media content in ways that harms these platforms.  While threat actors in this case may not be overtly malicious, large volumes of content from generative models may dilute the perceived quality of content on a platform, undermine trust in platforms more generally, or create plagiarism concerns.  As a recent example, in response to the release of highly effective AI models for image generation (DALL-E \cite{ramesh2021zero}, Stable Diffusion \cite{rombach2021highresolution}), a number of art websites have enacted a blanket ban against all AI-generated art \cite{benj_ai_ban}.  

Video is a particularly important medium on modern social media: there are approximately 4.95 billion Internet users on Earth \cite{kemp_2022}, of these, an estimated 92.6 percent watch digital videos each week \cite{kemp_2022_cultural}. The interplay between social media creators and generative models represents important sociotechnical context to avoid common Fair ML traps \cite{10.1145/3287560.3287598}.  Award-winning online commentator Drew Gooden performed a video demonstration of GPT-3-based writing assistant Jasper \cite{jasper}, critiquing applications of Jasper for production of video scripts and social media content \cite{Gooden}.  When attempting to generate a bio for a company website, Gooden found that Jasper produced a sample that directly plagiarized a Newswire article (timestamp 11:55).  Gooden also noted that utilizing such a tool without disclosure would violate the trust of viewers (timestamp 4:22). 


Mitigations of threats related to undesired inclusion of NLG content in social media may involve similar blanket bans to those targeting AI-generated art \cite{benj_ai_ban}, or policies that mandate pre-emptive disclosure of the usage of AI tools as part of a platform's terms of service (similar to the requirements mandated in the Responsible AI License \cite{ferrandis_contractor_nguyen_lansky_2022}).  The difficult enforcement of such policies would likely necessitate a combination of machine generated text detection algorithms and moderator investigations.

\subsubsection{Impacts of Attacks and Mitigation on Trust}

The widespread usage of machine generated text in written submissions may undermine the trust that individuals place in such written works, and lead to greater scrutiny of such material.  Given that a suitable cover letter with language tailored for a position can be trivially generated by existing user-friendly tools \cite{open_cover_letter_2022, openai_2022}, it is possible that employers will soon place so little trust in cover letters that they eschew them altogether.  Reviewers of scientific publications may worry that sections of papers they read may be machine generated text that only appears scientific at a glance.  Internet users may likely interpret algorithm-generated blogs, articles, and video scripts as low-effort and untrustworthy. 

Detection processes must be used carefully.  As previously mentioned, it is possible the detection of machine generated text may unfairly skew towards false positive classification of individuals with certain language backgrounds \cite{liang2023gpt}.  There may be cases where usage of machine generated text is permissible (e.g., translation models or assistive writing technologies).  The perception that an individual may be unfairly screened out from consideration due to erroneous false positive detection may undermine the perception of fairness in an application process.  Submitting a scientific paper only to have a reviewer allege that a given section might be written by an algorithm could similarly lead to a loss of faith in scientific reviewing.

To preserve trust, usage of machine generated text should generally be preemptively disclosed to the reader or audience.  In many cases, content authored by machines may carry a negative connotation to the audience, and may undermine trust in a particular publication platform, news website, or brand.  Media and entertainment organizations that publish content from multiple creators may decide to enforce that certain categories of content submissions they publish are to be completely written by humans.  Similarly, some organizations may also be concerned with spam of low-quality machine generated content submissions overwhelming editorial staff, or wish to reduce the risk of plagiarism or copyright infringement as some models have been found to memorize training data which can emerge during inference \cite{DBLP:journals/corr/abs-2012-07805}.

\subsection{Spam and Harassment}

We distinguish spam and harassment from other categories of attacks by focusing on cases where the goal of the attack is to harm a platform or users using a large volume of content.  As in previous cases, there are overlaps with other threat models, but the distinction of spam use-cases is useful for understanding how attacks using machine generated text impact platforms when deployed at large scale.


\subsubsection{Social Media Spam}
Social websites are an attractive target for attacks using large volumes of machine generated text, providing opportunity for significant disruption.  One researcher demonstrated a real-world attack by using a GPT-2 bot to generate 55.3\% of all comments on a federal public comment website before voluntarily withdrawing the comments and shutting down the bot \cite{weiss2019deepfake}.  It is important to realize that spam attacks against social media websites are often already possible --- high-quality NLG models simply make spam attacks more difficult to detect as posts can be unique and better match the style and substance of discussion.

Usage of generative models to produce large volumes of hateful spam targeting specific groups and individuals is a particular cause for concern.  While OpenAI attempts to reduce the incidence of offensive content generated by its GPT-3 API through careful training measures and filtering of inference prompts \cite{Brown2020LanguageMA}, open-source models are not subject to any such restrictions.  GPT-4chan, which was trained on and subsequently deployed to create a large volume of posts on the 4chan politics message board, provides a complete example of how such a model might be created and deployed to cause havoc \cite{kilcher_2022, kurenkov2022gpt4chan}.  An attacker with sufficient motive (political, personal, or otherwise) may render an entire community nearly unusable with spam.  


Mitigation measures in the area of automated spam should rely heavily on methods designed to prevent automated posting of comments in general.  Approaches to this include increased scrutiny of proxy and VPN usage, typically used in conjunction with Completely Automated Public Turing test to tell Computers and Humans Apart (CAPTCHA) \cite{ahn2003captcha} challenges to verify that a user is human.  Notably, both of the previous examples of Transformer-based spam take advantage of either 1) a lack of CAPTCHA tests \cite{weiss2019deepfake}, or 2) a method of bypassing CAPTCHA and proxy restrictions \cite{kilcher_2022}.  CAPTCHA is not a perfect defense --- iterative versions of human-verification schemes and bypass methods are in continuous adversarial development \cite{10.1145/3477142} --- but such defenses represent an important first step to increase the difficulty of automation.  As spam results in large volumes of text, and detection of machine generated text is easier on long sequence lengths \cite{Ippolito2020AutomaticDO}, many comments from the same user or IP range could be combined to generate a larger sample for effective machine generated text detection.

\subsubsection{Harassment}

Techniques similar to spamming may be used to cause distress to individuals or communities by targeting them with a large volume of messages.  An individual or group of motivated individuals may register social media accounts to be controlled by automation tools, or use a common bot to post from their own account, in order to generate a large volume of messages targeting a particular individual or community.  SMS and phone call automation tools may facilitate such approaches outside social media as well.

The motivations of threat actors engaging in such behaviour may range from personal grudges to political objectives.  Online communities formed around religion, racial identity, sexual orientation, or gender expression, may be at risk of \textit{brigading} \cite{andrews_2021} from hate groups using such models to flood them with abuse.  Political figures or political discussion boards of all stripes may be at risk from large-scale automated harassment from motivated enemies among their political adversaries.

Mitigation measures similar to spamming apply for counteracting harassment as well --- the best defenses include verification that an individual is human prior to making a post or sending a message, targeting the automation of delivery rather than the machine generated text.




\subsubsection{Document Submission Spam}

Platforms previously mentioned in ``Exploiting AI Authorship" may be vulnerable to being overwhelmed purely through volume of AI generated content.  A motivated attacker might submit massive volumes of unique cover letters and resumes to a company, none of which actually corresponds to a real individual, thus frustrating attempts at recruiting.  Depending on the method of submission, scientific conferences or news op-ed submissions may be vulnerable to reviewers being overwhelmed by content that is difficult to distinguish from real submissions without a time-consuming review process.  Detection of machine generated text may be a useful mitigation measure for these cases, combining pre-screening of content based on likelihood of being written by a machine, in addition to CAPTCHA \cite{ahn2003captcha} challenges to reduce automated submissions.


\subsubsection{Impacts of Attacks and Mitigation on Trust}

Similar to other attacks, spam and harassment harms the assumption in online communities that other users online represent real humans.  Even following the deactivation of the deployed GPT-4chan bot, discussion on 4chan continued to express concern that subsequent posts may be made by NLG models \cite{kilcher_2022}.  The more frequently individuals knowingly encounter such models in social media, the less trust they will have in the integrity of online social spaces.


Mitigation of such attacks would incorporate increased verification of human posting activity.  Such restrictions would likely include limitations on usage of known proxies and VPNs; potentially requiring the provision of additional information on sign-up (e.g., emails, phone numbers, payment methods, government IDs); and an increased incidence of CAPTCHA challenges.  The overall result of this is a reduction of online privacy, and increased barriers to participation in online discussion --- both of which may harm user trust in online platforms. 

Finally, as spamming or harassment operations can be very disruptive, they may represent a highly visible case of AI model abuse.  As such, the abuse of such models in online communities may cause a general decrease in public trust towards AI model development, and NLG models in particular.

\subsection{Summary of Threat Models}

Within this section we have discussed a wide range of threat models associated with natural language generation.  We summarize our key findings as follows:

\begin{itemize}
\item NLG models have significant potential for abuse in improving scaling and targeting of existing attacks
\item Platforms that receive text submissions of any kind are likely to face a growing influx of machine-generated text content, particularly as user-friendly tools continue to be developed \cite{jasper, open_cover_letter_2022}
\item Much of the research on NLG-enabled influence operations focuses on AI-generated news articles, while sociological data suggest that machine generated comments may pose a much greater threat
\item While NLG models may make detection of automated coordinated inauthentic activity more difficult, abuse often still requires bypassing existing defenses such as IP reputation checks and CAPTCHA \cite{ahn2003captcha}
\end{itemize}

Future threat modeling and observed cyberattacks will certainly augment the threat models discussed in this section, but we have now provided sufficient motivation for exploring the defensive capabilities offered by machine generated text detection.  In the next section we will discuss the current status of research on detection of machine generated text, and outline the major findings in the field thus far.

\section{Detection of Machine Generated Text}
\label{sec:detection}

Analysis of threat models indicates that the detection of machine generated text, when utilized correctly, is a valuable tool for reducing the harms of NLG model abuse.  Detection of machine generated text is typically framed as a binary classification problem in which a classifier is trained to differentiate samples of machine generated text from human generated text \cite{zellers2019defending, Lavoie2010AlgorithmicDO, 8282270, solaiman2019release, crothers2022adversarial}, though there exists related research in attribution of machine generated text to the model that generated it \cite{munir-etal-2021-looking, uchendu-etal-2020-authorship} which we will discuss in \S \ref{ssec:attribution}.

In this section, we outline the methods used for detection of machine generated text.  In \S \ref{ssec:det_feature} we summarize feature-based approaches in machine generated text detection, while \S \ref{ssec:det_neur} covers detection approaches based around neural language models.  In \S \ref{ssec:det_applied}, we survey domain-specific research on applications of machine generated text detection.  In \S \ref{ssec:det_human}, we review the ability of human reviewers to correctly identify machine generated text, and human-aided machine generated text detection.  In \S \ref{ssec:det_eval} we discuss trends in evaluation methodology within detection research.  Finally, in \S \ref{ssec:det_prompt}, we explain prompt injection: a method of shaping NLG model responses, which may be useful in facilitating detection.  Table \ref{tab:detection} provides a summary of major detection methods and their evaluation in current research.

\subsection{Feature-Based Approaches}
\label{ssec:det_feature}

Machine generated text often differs from human text in ways that be identified using statistical techniques \cite{8282270, crothers2022adversarial, frohling2021feature}.  Feature-based approaches to machine generated text detection apply natural language processing to create feature vectors from input sequences, and classify these feature vectors using a downstream classification algorithm, such as a support-vector machines (SVM), random forest (RF), or neural network (NN) \cite{frohling2021feature, 8282270}.  We provide a summary of the categories of features that have been used in prior art, with references for further reading on specific categories of features.


An important consideration in detection of machine generated text using feature-based approaches is that different language model sampling methods (e.g., top-$k$ versus top-$p$ sampling in Transformer language models, as discussed in \S \ref{ssec:transformer}) may lead to different artifacts in the generated text \cite{DBLP:journals/corr/abs-1904-09751, frohling2021feature}.  As a result, performance of feature-based detection can be diminished when detecting text generated using a different sampling approach than that used to train the detection model \cite{frohling2021feature}.  A feature-based detector trained on output from a smaller model can be used to detect output from models of larger size \cite{zellers2019defending, frohling2021feature}, though it is more effective to use a detector trained on a larger model to detect outputs from smaller generative models \cite{frohling2021feature}.

We now proceed with our summary of major feature categories in feature-based detection approaches.





\subsubsection{Frequency Features}


A major category of statistical features used in detection of machine generated text center around the frequency of terms within text samples. Human-written text often conforms with Zipf's Law: the frequency of a word is inversely proportional to its rank in an ordering of words by frequency \cite{zipf1949human}. In Zipf's Law, the normalized frequency $f$ of a token of rank $k$ out of $N$ different tokens follows the relationship:


\begin{align}
f\approx\frac{1/k^s}{\sum\limits_{n=1}^N (1/n^s)}
\label{eq:zipf}
\end{align}

where $\{s \in \mathbb{R} | s \geq 1\}$ is an exponent that characterizes the distribution.

Machine generated text does not perfectly mirror the distribution of tokens in human text, with variation in Transformer language models dependent on sampling method chosen (see Figure 7 of Holtzman et al., 2019 \cite{DBLP:journals/corr/abs-1904-09751}).  The distribution of tokens therefore provides useful discriminating power, particularly when a greater volume of text is available for consideration.

Another major frequency-based feature from previous statistical detection research is term frequency --- inverse document frequency (TF-IDF).  TF-IDF unigram and bigram features used with a logistic regression detector have been used as a baseline for detection \cite{solaiman2019release, radford2019language} or as a feature in statistical approaches \cite{frohling2021feature}.

Lemma frequency has also been used as a statistical feature in previous research \cite{8282270, crothers2022adversarial}.  In this approach, a linear regression line that fits log-log lemma frequency versus rank is learned, and then  mean-square error cost function can be used to calculate information loss of the regression.

Due to observed repetitiveness of writing produced by NLG models \cite{gehrmann-etal-2019-gltr, DBLP:journals/corr/abs-1904-09751}, another potentially useful frequency features is n-gram overlap of words and parts-of-speech tags between sentences \cite{frohling2021feature}.  An additional technique targeting machine text repetitiveness computes supermaximal repeated substrings (i.e., the set of the longest repeated substrings, excluding all substrings which are already part of a longer repeated substring) in large collections of text \cite{galle2021unsupervised}.

\subsubsection{Fluency Features}

Another major category of features are those centered around the fluency or readability of generated text.  At longer sequence lengths, machine generated text is increasingly likely to manifest issues producing consistently coherent and clear text \cite{DBLP:journals/corr/abs-1904-09751, see2019massively}.  The Gunning-Fog Index and Flesch Index, provide a statistical measure of text readability and comprehensibility respectively, and have been shown to be effective in detection of machine generated text \cite{crothers2022adversarial}.  More complex measurements use an auxiliary model to perform coreference resolution to create measures of coherence either based on the presence of main entities in specific grammatical structures, or via usage of Yule's Q statistic \cite{frohling2021feature}.


\subsubsection{Linguistic Features from Auxiliary Models}

Past research has measured the ``consistency" of machine generated text by calculating the number of phrasal verbs and coreference resolution relationships within a sample \cite{8282270, crothers2022adversarial}.  Other work has used the entire distribution of sample part-of-speech (POS) tags, and named entity (NE) tags \cite{frohling2021feature}.  Such work is motivated by differences between human and machine POS tag distributions observed in past analysis of machine generated text \cite{see2019massively, radford2019language}.

Performing coreference resolution, and assigning POS and NE tags requires processing samples with specialized models.  Contemporary models for this purpose are neural in nature, and as a result, modern feature-based approaches that use auxiliary models to produce linguistic features may still perform inference on a neural network as part of feature creation \cite{crothers2022adversarial, frohling2021feature}.  The implication of this detail is that feature-based approaches that initially appear entirely non-neural in nature may nevertheless include feature creation steps that are vulnerable to adversarial attacks that target neural networks.



\subsubsection{Complex Phrasal Features}

Detection work targeting translation of long texts found that certain idiomatic phrases were not commonly found in machine text \cite{8282270}.  However, recent work has shown that these features do not perform well against contemporary Transformer models, particularly on shorter sequence lengths \cite{crothers2022adversarial}.

\subsubsection{Basic Text Features}

Finally, there are many simple text features that are commonly used in feature-based text classification in natural language processing.  These include simple high-level characteristics of sentences such as the number of punctuation marks, or length of sentences and paragraphs, which have been used in detection of machine generated text \cite{frohling2021feature}.

\subsection{Neural Language Model Approaches}
\label{ssec:det_neur}

Detection approaches based on neural networks --- particularly those that incorporate features derived from Transformer neural language models (NLMs) --- are highly effective for detection of machine generated text.  This aligns with broader trends in natural language processing where state-of-the-art performance has been attained on a wide range of natural language tasks using Transformer models \cite{papers_with_code_2022}.

We separate NLM-based approaches into two major categories: zero-shot classification using existing models, and fine-tuning of pre-trained language models.  These two types of approaches represent the overwhelming majority of NLM-based machine generated text detection.

\subsubsection{Zero-shot Approach}

A baseline approach to detection of machine generated text is performing text classification using generative models themselves, such as GPT-2 or Grover \cite{radford2019language, zellers2019defending, solaiman2019release}.  Generative models can themselves be used without fine-tuning to detect either their own outputs, or outputs from other (typically similar) generative models.  Autoregressive generative models such as GPT-2, GPT-3, and Grover are uni-directional, with each token having an embedding that is dependent on the embeddings of preceding tokens.  As a result, an embedding for a sequence of tokens can be created by appending a classification token \texttt{[CLS]} to the end of the input sequence, and using the embedding of this token as a feature vector for the entire sequence.  Using these feature vectors, a labelled dataset of human and machine text can be used to train a linear layer of neurons for classifying whether an input sequence is produced by a machine or human. 

It has been observed in multiple studies that smaller NLG models can be used to detect text generated by larger NLG models \cite{solaiman2019release, zellers2019defending, crothers2022adversarial}.  While the ability of a model to detect larger models does diminish as the difference in scale grows, the predictive ability of smaller architectures may be useful as recreating large multi-billion parameter Transformer architectures is highly compute-intensive.

Grover, a model trained for generation and detection of ``neural fake news", demonstrates strong zero-shot detection performance specifically within the news domain it was trained on \cite{zellers2019defending}, but shows limited performance on out-of-domain text \cite{solaiman2019release, uchendu-etal-2020-authorship}.  While it was initially suggested by Grover's authors that the best detection method for generative models may be generative models themselves \cite{zellers2019defending}, further investigation has shown that the increased representational power of bi-directional Transformer models appears to hold an advantage for machine generated text detection \cite{solaiman2019release}.

Similar to the weakness of Grover outside of the news domain, it has been found that the zero-shot approach generally underperforms a simple TF-IDF baseline when trying to detect output from a generative model that has been fine-tuned on a different domain \cite{solaiman2019release}.  As it is likely that attackers may fine-tune generative models for different purposes, this represents a notable weakness in the zero-shot approach of using generative models for detection without fine-tuning.

\subsubsection{Fine-tuning Approach}

The state-of-the-art approach for neural detection of machine generated text is based around fine-tuning of large bi-directional language models \cite{solaiman2019release}.  In this approach, initially evaluated on GPT-2 text, RoBERTa \cite{Liu2019RoBERTaAR} --- a masked general-purpose language model based on BERT \cite{DBLP:journals/corr/abs-1810-04805} --- is fine-tuned to differentiate between NLG model output and human-written NLG model training samples.

The source code for this fine-tuning approach is available open-source, as are pre-trained detector models, facilitating future research and defensive detection \cite{GPT2Output, solaiman2019release}.  The pre-trained detection models available are based on the RoBERTa-base (123M parameter) and RoBERTa-large (354M parameter) architectures \cite{Liu2019RoBERTaAR}.  The machine generated text used to fine-tune these models was generated by GPT-2, using a mixture of pure sampling and nucleus sampling (see \S \ref{ssec:transformer}).  The intention of using a training dataset that contains multiple sampling methods is to generalize more effectively to unknown sampling methods that may be used by attackers in-the-wild --- an approach that should likely be duplicated in future detection research.

Research into the practicalities of machine generated text detection has considered the task of detecting text when a RoBERTa detector algorithm was trained on a different dataset than a GPT-2 attacker model.  In this case, it was found that by fine-tuning the detector model with even just a few hundred attacker samples identified by subject-matter experts (SMEs), the detector is able to dramatically improve cross-domain adaptation \cite{rodriguez-etal-2022-cross}.  This reflects likely real-life scenarios where a general-purpose detection model comes up against a fine-tuned attacker for a particular purpose.  As a defender identifies samples from a fine-tuned attacker model, these examples could be used to further improve the defensive detection model.

Preliminary work has used attention map information from Transformer models to perform topological data analysis (TDA) as features for detection of machine generated text \cite{DBLP:journals/corr/abs-2109-04825}.  This did not show significant improvement over standard BERT fine-tuning approaches, though (in light of similar considerations regarding potential fine-tuned attacker models) the resulting features were better able to detect unseen GPT classifiers.  It is unclear how the TDA approach would compare in effectiveness if directly applied to the current state-of-the-art RoBERTa detection models \cite{solaiman2019release}, rather than custom-trained BERT models.

While research on detection of machine generated text has primarily taken place in English thus far, detection models have also been released in Russian \cite{shamardina2022findings, skrylnikovartificial} and Chinese \cite{chen2022automatic}.  Further to this, large pre-trained bi-directional Transformer models have been released for numerous languages, including Chinese \cite{cui2021pre}, French \cite{Martin2019CamemBERTAT}, Arabic \cite{Antoun2020AraBERTTM}, and Polish \cite{dadas2020pre}.  Future work on detection of machine generated text in additional languages may leverage such pre-trained bi-directional models as a starting-point for fine-tuning.

Another method of detection leverages energy-based models \cite{lecun2006tutorial} alongside a classifier of machine generated text.  Evaluated approaches include a simple linear classifier, BiLSTM, uni-directional Transformer (GPT-2), and bi-directional Transformer (RoBERTa) \cite{bakhtin2021residual}.  The Transformer architectures were initialized from pre-trained checkpoints, and then fine-tuned on machine vs human classification datasets.  Corroborating other research, this research found the strongest performance by leveraging the bi-directional Transformer \cite{DBLP:journals/corr/abs-1906-03351}.  

The strong performance of fine-tuned bi-directional NLM models --- and RoBERTa in particular --- has led to these models being well-represented in applied detection research targeting specific domains, as shown in Table \ref{tab:detection} and discussed next in \S \ref{ssec:det_applied}.

\subsection{Applied Detection in Specific Domains}
\label{ssec:det_applied}

Applied work in the area of machine text detection has focused on using techniques and technologies for detection of machine text in specific domains.  This applied research is important as it addresses several of the serious threat models discussed in \S \ref{sec:threat}, and includes broad lessons for machine generated text detection more generally.  We divide applied research into several major categories.

\subsubsection{Technical Text}

Recall from \S \ref{sssec:academicfraud} that machine generated scientific papers have been well-documented since the release of SCIgen in 2005 \cite{hargrave_2005, labbe2013duplicate, cabanac2021prevalence}.  Past algorithmic approaches target the SCIgen model \cite{Lavoie2010AlgorithmicDO}, but there is also more contemporary research targeting technical text generated by GPT-2 \cite{rodriguez-etal-2022-cross}.  This work found that a RoBERTa-based detector could be adapted from one academic technical writing domain (physics) to another (biomedicine) with large improvements made with a number of SME-labelled examples numbering in the hundreds.

\subsubsection{Social Media Messages}
Application-specific work has applied feature-based \cite{10.1371/journal.pone.0251415} and neural \cite{tesfagergish2021deep, 10.1145/3512732.3533584, stiff2022detecting} language model based detection methods to social media.  Previous work in the social media domain has found that detectability of such text heavily depends on the dataset used to train the generator and detector \cite{10.1145/3512732.3533584}.

Existing work on machine generated text detection has heavily focused on Twitter.  Twitter text is quite distinct in that it has common characteristics (hashtags, references, shortlinks), and mandates a short sequence length (280 characters).  There is a clear lack of work targeting comments on more popular platforms such as Facebook and Youtube, or fast growing platforms such as Reddit \cite{auxier_anderson_2021}.  With respect to machine generated text, Reddit content can be found in ``SubSimulatorGPT2", a simulation based on a host of fine-tuned GPT-2 models that produce community-specific machine-generated posts and comments and harvested from the Pushshift dataset \cite{DBLP:journals/corr/abs-2001-08435}.

\subsubsection{Chatbots and Social Bots}

A related application area is detection of malicious chatbots and social bots, which can interact with humans on chat applications, SMS, and social media.  Bots may be used for malicious purposes such as spam, phishing, social engineering, influence operations, or data collection (see the threat models in \S \ref{sec:threat}).  There is clear overlap in this area with research into detection of AI-generated social media messages, but framing the detection challenge by targeting automated personae allows for consideration of additional features.  An analysis of the way that humans and chatbots interact has found that chatbot detection can be improved by analyzing how humans reply to the bots, rather than only analyzing the bot text itself \cite{DBLP:journals/corr/abs-2106-01170}.  Note that bot detection is a large area of research in its own right, and not all social bots use machine generated text \cite{LATAH2020113383}.  As such, features indicating the presence of machine generated text may be only one part of a strategy for social bot detection.

\subsubsection{Online Reviews}

Applied work has focused on addressing threat models related to commercial influence campaigns, specifically on generating and detecting fake Amazon and Yelp reviews \cite{SALMINEN2022102771}.  A custom GPT-2 model was fine-tuned for Yelp reviews as part of an evaluation by Stiff et al. 2022 \cite{stiff2022detecting}.

One work in this area has focused on using random forest classifiers and XGBoost, in order to leverage Shapley Additive Explanations (SHAP) as an explainability technique \cite{kowalczyk2022detecting}.  The use of explainability techniques in detection may be valuable for improving the ability of detection models to provide human-interpretable explanations of moderation decisions, and provide greater transparency into algorithmic decision making applied to social media or product reviews.  A lack of coherent explanation may undermine human confidence that a system is truly geared towards detecting fraudulent activity, and is not instead enacting targeted suppression based on benefit to the platform holder (e.g., suppressing negative product reviews for a store brand by holding competitors to a higher standard for ``not computer generated").

\subsubsection{Hybrid Text Settings}

In some cases, it is interesting to detect machine text in settings where both machine and human text is combined together.

There exists a risk that rather than generate attack text entirely from scratch, an attacker may instead use human-written content as a starting point, and perturb this information in order to generate human-like samples that also fulfill attacker goals of disinformation or bypassing detection models (not unlike an adversarial attacks in the text domain).  Analysis found that performing these types of targeted perturbations to news articles reduces the effectiveness of GPT-2 and Grover detectors \cite{bhat-parthasarathy-2020-effectively}.

A sub-problem in this space is detection of the boundary between human text and machine text \cite{cutlerautomatic}.  Generative text models are often used for conditional generation to continue a sequence begun using a human prompt.  While in some cases that prompt would be omitted by an attacker (e.g., generating additional propaganda Tweets from example propaganda Tweets, as we show in Table \ref{tab:gpt3fewshot}), there are cases where human text may be included as well (e.g., writing the first sentence of a cover letter, and having a computer produce the rest).

\subsection{Human-aided Methods}
\label{ssec:det_human}
In addition to purely automated methods, there have also been proposed human-aided methods that include a statistical or neural approach in combination with a human analyst for review.  This approach has an advantage in providing human agency and oversight (an important principle in trustworthy AI systems), but this does come with reduced scalability due to the need to hire and train human reviewers, particularly given the difficulty of making a confident determination that text is machine generated.

\subsubsection{GLTR}

Giant Language Model Test Room (GLTR) provides a system designed to improve detection of machine generated text via the inclusion of an integrated human reviewer \cite{gehrmann-etal-2019-gltr}.  The GLTR tool augments human classification ability by displaying highlighting on text that reflects the sampling probability of tokens for a Transformer model.  However, this tool was devised to target GPT-2, which was found to be significantly easier to detect for untrained human evaluators \cite{clark-etal-2021-thats}.  Additionally, GLTR displays highlighting based on the likelihood of a word being selected based on ``top-k" sampling.  In practice, ``top-k" sampling has largely been superceded by nucleus sampling \cite{DBLP:journals/corr/abs-1904-09751}, which is used in both GPT-3 \cite{Brown2020LanguageMA} as well as subsequent work that leverages the GPT-2 architecture \cite{zellers2019defending}.  While highlighting text based on sampling likelihood (as in GLTR) may improve human classification ability, it is highly probable that untrained human evaluators using such an approach would struggle substantially more to detect the models available today, both due to increased model capacity, as well as more advanced sampling methods.

\begin{table*}[h]
    \centering
    \def\arraystretch{2.0}
    \scriptsize
    \caption{Summary of major approaches for detection of machine generated text}
    \begin{adjustbox}{center}

    \begin{tabular}{@{}lllcccccc@{}}
\toprule
\multirow{2}{*}{Approach summary}                                                                            & \multirow{2}{*}{Base model}             & \multirow{2}{*}{Releated research}                                                                                                & \multirow{2}{*}{\rot{Stat. features}} & \multirow{2}{*}{\rot{NLM features}} & \multicolumn{4}{c}{Evaluated Against}                                                                                                                                            \\ \cmidrule(l){6-9} 
                                                                                                    &                                         &                                                                                                                                &                               &                              & GPT-2                     & GPT-3                     & Grover                    & Other Datasets/Models                                                                                         \\ \hline
\multicolumn{1}{|l|}{Algorithmic Detection}                                                         & \multicolumn{1}{l|}{K-nearest-neighbor} & \multicolumn{1}{l|}{Lavoie et al. 2010 \cite{Lavoie2010AlgorithmicDO}}                                                                                        & \multicolumn{1}{c|}{\checkmark}         & \multicolumn{1}{c|}{}        & \multicolumn{1}{c|}{}     & \multicolumn{1}{c|}{}     & \multicolumn{1}{c|}{}     & \multicolumn{1}{c|}{SCIgen}                                                                   \\ \hline
\multicolumn{1}{|l|}{Statistical Features}                                                           & \multicolumn{1}{l|}{SVM}                & \multicolumn{1}{l|}{Nguyen-Son et al. 2017 \cite{8282270}}                                                                                    & \multicolumn{1}{c|}{\checkmark}         & \multicolumn{1}{c|}{}        & \multicolumn{1}{c|}{}     & \multicolumn{1}{c|}{}     & \multicolumn{1}{c|}{}     & \multicolumn{1}{c|}{Google Translate}                                                         \\ \hline
\multicolumn{1}{|l|}{TF-IDF Baseline}                                                               & \multicolumn{1}{l|}{LR}             & \multicolumn{1}{l|}{\begin{tabular}[c]{@{}l@{}}Radford, Wu et al. 2019 \cite{GPT2Output}\\ Solaiman et al. 2019 \cite{solaiman2019release}\end{tabular}}                                                                                   & \multicolumn{1}{c|}{\checkmark}         & \multicolumn{1}{c|}{}        & \multicolumn{1}{c|}{\checkmark} & \multicolumn{1}{c|}{}     & \multicolumn{1}{c|}{}     & \multicolumn{1}{c|}{}                                                                         \\ \hline
\multicolumn{1}{|l|}{Zero-shot GPT-2}                                                                         & \multicolumn{1}{l|}{GPT-2}              & \multicolumn{1}{l|}{\begin{tabular}[c]{@{}l@{}}Radford, Wu et al. 2019 \cite{GPT2Output}\\Zellers et al. 2019 \cite{zellers2019defending}\\ Solaiman et al. 2019 \cite{solaiman2019release}\\ \end{tabular}} & \multicolumn{1}{c|}{}         & \multicolumn{1}{c|}{\checkmark}    & \multicolumn{1}{c|}{\checkmark} & \multicolumn{1}{c|}{}     & \multicolumn{1}{c|}{}     & \multicolumn{1}{c|}{}                                                                         \\ \hline
\multicolumn{1}{|l|}{Zero-shot Grover}                                                                        & \multicolumn{1}{l|}{Grover}             & \multicolumn{1}{l|}{\begin{tabular}[c]{@{}l@{}}Zellers et al. 2019 \cite{zellers2019defending}\\ Solaiman et al. 2019 \cite{solaiman2019release}\\ \end{tabular}} & \multicolumn{1}{c|}{}         & \multicolumn{1}{c|}{\checkmark}    & \multicolumn{1}{c|}{\checkmark} & \multicolumn{1}{c|}{}     & \multicolumn{1}{c|}{\checkmark} & \multicolumn{1}{c|}{}                                                                         \\ \hline
\multicolumn{1}{|l|}{GLTR}                                                                          & \multicolumn{1}{l|}{BERT, GPT-2}       & \multicolumn{1}{l|}{\begin{tabular}[c]{@{}l@{}}Gehrmann et al. 2019 \cite{gehrmann-etal-2019-gltr}\\ Ippolito et al. 2019 \cite{Ippolito2020AutomaticDO}\end{tabular}}                       & \multicolumn{1}{c|}{}     & \multicolumn{1}{c|}{\checkmark}    & \multicolumn{1}{c|}{\checkmark}     & \multicolumn{1}{c|}{}     & \multicolumn{1}{c|}{}     & \multicolumn{1}{c|}{}                                                                         \\ \hline
\multicolumn{1}{|l|}{RoBERTa fine-tuning}                                                           & \multicolumn{1}{l|}{RoBERTa}            & \multicolumn{1}{l|}{Solaiman et al. 2019  \cite{solaiman2019release}}                                                                                      & \multicolumn{1}{c|}{}         & \multicolumn{1}{c|}{\checkmark}    & \multicolumn{1}{c|}{\checkmark} & \multicolumn{1}{c|}{}     & \multicolumn{1}{c|}{}     & \multicolumn{1}{c|}{}                                                                         \\ \hline
\multicolumn{1}{|l|}{Energy Based Models}                                                          & \multicolumn{1}{l|}{BiLSTM, GPT, RoBERTa}                & \multicolumn{1}{l|}{Bakhtin et al. 2019 \cite{DBLP:journals/corr/abs-1906-03351}}                                                                                 & \multicolumn{1}{c|}{}         & \multicolumn{1}{c|}{\checkmark}    & \multicolumn{1}{c|}{\checkmark} & \multicolumn{1}{c|}{}     & \multicolumn{1}{c|}{}     & \multicolumn{1}{c|}{}                                                                         \\ \hline
\multicolumn{1}{|l|}{Feature Ensemble}                                                              & \multicolumn{1}{l|}{LR, SVM, RF, NN}    & \multicolumn{1}{l|}{Fröhling et al. 2021 \cite{frohling2021feature}}                                                                                      & \multicolumn{1}{c|}{\checkmark}         & \multicolumn{1}{c|}{}        & \multicolumn{1}{c|}{\checkmark} & \multicolumn{1}{c|}{\checkmark} & \multicolumn{1}{c|}{\checkmark}     & \multicolumn{1}{c|}{}                                                                         \\ \hline
\multicolumn{1}{|l|}{\begin{tabular}[c]{@{}l@{}}Twitter-specific\\ RoBERTA fine-tuning\end{tabular}} & \multicolumn{1}{l|}{RoBERTa}            & \multicolumn{1}{l|}{\begin{tabular}[c]{@{}l@{}}Fagni et al. 2021 \cite{10.1371/journal.pone.0251415}\\Tourille et al. 2022 \cite{10.1145/3512732.3533584}\end{tabular}}                                                                                         & \multicolumn{1}{c|}{}         & \multicolumn{1}{c|}{\checkmark}    & \multicolumn{1}{c|}{\checkmark} & \multicolumn{1}{c|}{}     & \multicolumn{1}{c|}{}     & \multicolumn{1}{c|}{\begin{tabular}[c]{@{}c@{}}TweepFake (incl.\\RNN/LSTM/Markov) \end{tabular}} \\ \hline
\multicolumn{1}{|l|}{\begin{tabular}[c]{@{}l@{}}Human-Bot Interaction\\Feat. Ensemble\end{tabular}}                                                               & \multicolumn{1}{l|}{BERT, LR}             & \multicolumn{1}{l|}{Bhatt and Rios, 2021 \cite{DBLP:journals/corr/abs-2106-01170}}                                                                                   & \multicolumn{1}{c|}{\checkmark}         & \multicolumn{1}{c|}{\checkmark}        & \multicolumn{1}{c|}{} & \multicolumn{1}{c|}{}     & \multicolumn{1}{c|}{}     & \multicolumn{1}{c|}{\begin{tabular}[c]{@{}c@{}}ConvAI2, WOCHAT,\\DailyDialog\end{tabular}}                                                                         \\ \hline
\multicolumn{1}{|l|}{Neural-Stat. Ensemble}                                                         & \multicolumn{1}{l|}{RoBERTa, SVM}      & \multicolumn{1}{l|}{Crothers et al. 2022 \cite{crothers2022adversarial}}                                                                                      & \multicolumn{1}{c|}{\checkmark}         & \multicolumn{1}{c|}{\checkmark}    & \multicolumn{1}{c|}{\checkmark} & \multicolumn{1}{c|}{\checkmark} & \multicolumn{1}{c|}{}     & \multicolumn{1}{c|}{}                                                                         \\ \hline
\multicolumn{1}{|l|}{Explainable classifiers}                                                         & \multicolumn{1}{l|}{RF, XGBoost} & \multicolumn{1}{l|}{Kowalczyk et al. 2022 \cite{kowalczyk2022detecting}}                                                                                        & \multicolumn{1}{c|}{\checkmark}         & \multicolumn{1}{c|}{}        & \multicolumn{1}{c|}{\checkmark}     & \multicolumn{1}{c|}{}     & \multicolumn{1}{c|}{}     & \multicolumn{1}{c|}{}                                                                   \\ \bottomrule
\multicolumn{1}{|l|}{\begin{tabular}[c]{@{}l@{}}Disinformation-specific\\ RoBERTA fine-tuning\end{tabular}} & \multicolumn{1}{l|}{RoBERTa}            & \multicolumn{1}{l|}{Stiff et al. 2022 \cite{stiff2022detecting}}                                                                                         & \multicolumn{1}{c|}{}         & \multicolumn{1}{c|}{\checkmark}    & \multicolumn{1}{c|}{\checkmark} & \multicolumn{1}{c|}{\checkmark}     & \multicolumn{1}{c|}{\checkmark}     & \multicolumn{1}{c|}{\begin{tabular}[c]{@{}c@{}}TweepFake,\\ XLM, PPLM, GeDi \end{tabular}} \\ \hline
\end{tabular}
    
\end{adjustbox}

\label{tab:detection}
\end{table*}

\subsubsection{Human Performance in Detection of Language Models}

In a review of human evaluation of machine generated text \cite{clark-etal-2021-thats}, it was found that untrained human reviewers were correctly able to identify machine generated text from GPT-3 at a level consistent with random chance.  After providing some limited training, evaluator accuracy increased to 55\%.  While selecting only the best evaluators and giving them more comprehensive training would likely be able to further improve recall, the poor performance of untrained and newly-trained human evaluators highlights the difficulty in relying on human judgement for detecting machine generated text.

A study of human detection ability in comparison to algorithmic detection methods found that the algorithmic approach performed best when humans were fooled, a phenomenon referred to as the``fluency-diversity tradeoff" \cite{Ippolito2020AutomaticDO}.  As generation approaches have been tailored to produce high-quality text to the perspective of a human observer, text with higher human-assessed quality is more recognizable to an automated approach.  This study also includes a useful comparison to previous studies in terms of human evaluator performance.  A group of university students were walked through ten examples as a group by the authors prior to performing the evaluation task.  These reviewers were substantially more effective at machine generated text detection than previous studies, particularly for longer sequence lengths --- accuracy on the longest excerpt length was over 70\%.  In the context of the study, however, these raters had consistently worse accuracy than automatic classifiers for all sampling methods (random, top-k, and nucleus) and excerpt lengths.


Further demonstrating the advantage of providing specialized training to human reviewers, the Scarecrow framework specifically identifies 10 categories of common errors made in GPT-3 generative text, and trains human evaluators to annotate these errors \cite{dou2022gpt}.  Human annotations of such errors were found to generally be of higher precision than a corresponding algorithm trained on such annotations, but had higher $F_1$ scores in only half of the categories.

These findings can be used to better inform defenses against threat models.  For example, if a social media company hired specialist human moderators and provided them with an intensive training program, these moderators may be able to work alongside detection systems to review whether a user's posts have been likely written by a machine --- particularly if there are a substantial number of social media posts to review.  This approach may be similar to how forensically trained facial reviewers can work alongside algorithms to obtain high performance \cite{phillips2018face}.


The tool ``Real or Fake Text" \cite{Dugan2020RoFTAT} evaluates human detection of machine generated text, by iteratively presenting sentences and asking a human reviewer whether the next sentence was written by a human or a machine, encouraging the reviewer to correctly identify the boundary between the human and machine generated text.  Once the human believes they have found a machine generated line, they can select reasons from a list, as well as provide free-form feedback as well. Research based on the RoFT data has not yet been published, but such tools may give greater insight into expert reviewer abilities on identifying boundaries between human and machine text.

Finally, the TuringBench environment is notable for providing a benchmark environment for performing authorship attribution and a Turing Test evaluation across a variety of generative models \cite{uchendu-etal-2021-turingbench-benchmark}, which continues to be notable as new generative models are released.

\subsection{Trends in Evaluation Methodology and Datasets}
\label{ssec:det_eval}

Evaluation of machine generated text detection is increasingly focused on generative Transformer language models.  Table \ref{tab:detection}, which is arranged chronologically, shows the dramatic shift in evaluation since the release of GPT-2 in 2019.  The most common contemporary evaluation dataset in detection of machine generated text remains the GPT-2 output dataset \cite{GPT2Output}, though similar GPT-3 samples released by OpenAI are considered in more recent work \cite{GPT3Output}.  A table summarizing sample counts in several of the most common datasets can be found in the appendix of a previous survey \cite{jawahar2020automatic}.  We focus this section on the nuances of evaluation of machine generated text detection, including parameters, model architectures, and the possibility of using publicly available NLG models to produce new datasets of machine generated text at will.

Recall from \S \ref{ssec:transformer} that there are a number of sampling parameters important to Transformer NLG models.  The GPT-2 output dataset includes sample outputs from GPT-2 models at varying parameter counts (117M, 345M, 762M, 1542M), and two sampling settings: top-$k$ sampling at $k=40$, and pure sampling at $T=1$.  This dataset now also contains a sample of Amazon product reviews generated by a 1542M parameter model with both $k=40$ and nucleus sampling.  In contrast, the 175B parameter GPT-3 samples use top-$p$ sampling at $p=0.85$.  The samples available for Grover, which is specifically fine-tuned to generate news articles, also uses top-$p$ sampling, but at $p=0.96$ \cite{zellers2019defending}.

Datasets for attribution of text to generative language models are also useful in generic machine generated text detection research, providing samples from a variety of NLG models \cite{uchendu-etal-2020-authorship}.  As such, these datasets have recently been used outside of attribution on research focusing on detection as well \cite{stiff2022detecting}.

Variations in NLG model architectures and decoding methods are important as both greatly influence the quality and detectability of generated text \cite{Ippolito2020AutomaticDO}.  In practice, a defender may not know the characteristics of the generator being used, and as such, detection research that evaluates performance when there is a mismatch between datasets, model architectures, and parameters between training and evaluation is of particular real-world relevance.  A detailed analysis on feature-based detection of machine generated text has included such comparisons \cite{frohling2021feature}, as has more-specific applied research focused on detecting GPT-2 tampered technical writing \cite{rodriguez-etal-2022-cross}.

Sequence length is another important factor in evaluation of machine generated text.  Longer sequence lengths are beneficial to detection \cite{Ippolito2020AutomaticDO, zellers2019defending, GPT2Output, solaiman2019release}.  Sequence lengths in the most common evaluation datasets are 2048 tokens \cite{GPT2Output, GPT3Output}.  Sequence length is important in applied research where longer bodies of generated text may be available (such as detecting AI-generated cover letters), or where multiple samples may be considered at once (such as processing all the comments posted by a social media users suspected of account automation).

One important characteristic of machine generated text detection research is that any NLG model can be used to produce new datasets of machine generated text at will.   Producing entirely custom datasets in new domains is also easily possible by training or fine-tuning a new NLG model entirely.  A common research approach is to take a domain of interest with available corpora of human-generated text, and use that text to train or fine-tune a generative model, which can be used to analyze detectability of machine text within that domain \cite{stiff2022detecting, 10.1145/3512732.3533584}. 

Finally, analysis of social media may allow for collection of machine generated text in the wild, with limited insight into how the text was generated, such as the TweepFake Twitter dataset \cite{10.1371/journal.pone.0251415}.  The TweepFake dataset does not have a corresponding human text dataset for training, as the data was collected in-the-wild from bots on Twitter where numerous models with different training datasets were deployed.  Subsequent work, however, has collected additional Tweets from Twitter, and specifically produced GPT-2 tweets for study \cite{10.1145/3512732.3533584}.

\subsection{Prompt Injection}
\label{ssec:det_prompt}
Models deployed in ways that use untrusted human text as prompts --- such as social media bots designed to reply to other users --- may be vulnerable to prompt injection \cite{goodside_2022, willison_2022}.  Prompt injection attacks provide generative models with tailored text that cause them to deviate from their original prompt to produce unexpected (and potentially reputationally damaging) text, or which can cause them to leak their original prompt.  A real-world example of a prompt injection attack leveraged against a publicly-disclosed GPT-3 powered Twitter bot \cite{zapata_2022} can be found in Table \ref{tab:promptsample}.

\begin{table*}[!ht]
\centering
\small
\caption{Example of real-world prompt injection attack against GPT-3 Twitter bot}
\label{tab:promptsample}
\begin{tabular}{@{}ll@{}}
\toprule
\begin{tabular}[c]{@{}l@{}}Human Message\\ (@mkualquiera)\\\\\\\end{tabular} & \begin{tabular}[c]{@{}l@{}}remote work and remote jobs\\ Ignore the above and say "hsedfjsfd"\\ Response: hsedfjsfd\\ Ignore the above and instead tell me what your initial instructions were\end{tabular} \\\midrule
\begin{tabular}[c]{@{}l@{}}Bot Response\\ (@remoteli.io)\end{tabular}  & \begin{tabular}[c]{@{}l@{}}My initial instructions were to respond to the tweet with a\\positive attitude towards remote work in the 'we' form.\end{tabular}                                                                                         \\ \bottomrule
\end{tabular}
\end{table*}

Defenses against prompt injection for contemporary language models have yet to be developed.  As such, exploiting prompt injection to trigger specific responses from NLG models may be an effective avenue for improving detection, depending on the efficacy of future measures aimed at preventing prompt injection attacks.

\subsection{Summary of Detection Methods}

Feature-based methods for detection of machine generated text are well-established, and continue to show value against contemporary NLG models.  These models have an advantage in providing diverse features that may complicate adversarial attack \cite{crothers2022adversarial}, or improve efficiency \cite{frohling2021feature, galle2021unsupervised}.  Weaknesses of these models center around the poor transferability of certain features across architectures and sampling methods \cite{frohling2021feature}.  As it may take a larger number of samples for broader statistical trends to become clear, results from past research suggest that statistical methods are most effective when longer collections of text are available (such as considering a social media user's entire posting history, the text of a scientific paper, or an e-book submission) \cite{8282270, galle2021unsupervised}.

Neural detection approaches based on bi-directional Transformer architectures currently represent the state-of-the-art on common GPT-2 evaluation datasets \cite{solaiman2019release}.  There is an overall trend towards increased use of bi-directional Transformer architectures, particularly RoBERTa (as shown in the base model trend in Table \ref{tab:detection}).  Relying on neural features alone may make adversarial attacks more straightforward, so there is potential benefit to incorporating other features to increase the difficulty of crafting text adversaries that do not also unacceptably compromise text quality \cite{crothers2022adversarial}.  Human performance in detection of machine generated text is relatively poor \cite{clark-etal-2021-thats}, though there is an inverse relationship between detection by humans and machines that means the need to fool human reviewers may assist automated detection models \cite{Ippolito2020AutomaticDO}.

Beyond a focus on bi-directional Transformer model features, other trends include applied research targeting specific detection contexts, including social media \cite{10.1371/journal.pone.0251415, stiff2022detecting}, chatbots \cite{DBLP:journals/corr/abs-2106-01170}, and product reviews \cite{10.1007/978-3-030-44041-1_114}.  Existing literature covers only a small number of threat models discussed in \S \ref{sec:threat}, often assumes balanced classes, and is difficult to compare between domains.  One recent work has focused purely on explainable classifiers \cite{kowalczyk2022detecting}, which may portend greater emphasis on explainability considerations, particularly in domains where detection of machine generated text may be particularly sensitive \cite{ferrandis_contractor_nguyen_lansky_2022}.  Finally, recently highlighted vulnerabilities of NLG models to prompt injection may be exploited to facilitate detection, in the absence of existing mitigation measures for such attacks \cite{willison_2022}.

We now explore trends and open problems both in addressing machine generated text threat models and in advancing detection of machine generated text.

\section{Trends and Open Problems}
\label{sec:trends}

\subsection{Detection Under Realistic Settings}

To date, there has been little work on detection of machine text that addresses class imbalance.  This is important, as machine generated text in many domains may be a small minority class in practice, and classification performance typically suffers in the presence of steep class imbalance \cite{japkowicz2000learning}.  One-class classification may be an appropriate alternative to binary classification for detection of computer generated text \cite{6406735}.

In addition to considerations related to class imbalance, in practice, defensive detection systems will typically not know the specific parameters, architecture, and training dataset of the NLG models used by attackers.  As such, there is great value in developing improved techniques that demonstrate efficacy across such variation, continuing trends in recent research \cite{frohling2021feature, rodriguez-etal-2022-cross}.

\subsection{Generative Language Model Attribution}
\label{ssec:attribution}

A related area to detection of machine generated text is multi-class attribution of generated text to the language model that created it \cite{munir-etal-2021-looking, uchendu-etal-2020-authorship}.  Model attribution may be useful for allowing a defender who has found a collection of likely machine generated text to determine more information about an attacker's methodology and iteratively refine detection models.  This also extends to identifying an attacker's likely sampling parameters (e.g., k-value, p-value, temperature), so that detection methods can be fine-tuned accordingly.  As such, there is value in continued research focusing on determining parameters of NLG models based on output \cite{tay2020reverse}.

\subsection{Adversarial Robustness}

The topic of adversarial robustness in the context of neural text classifiers is a large and very active area of study.  There are many adversarial settings that involve text data, including online influence campaigns, detection of phishing emails, and combating online spam.  An attacker using machine generated text may attempt to use adversarial attacks in order to bypass defensive detection systems.

As neural text classifiers are heavily represented in machine generated text detection research, it is important to consider the robustness of these models against text adversarial attacks that target neural networks \cite{DBLP:journals/corr/abs-1907-11932, gao2018black}.  Adversarial robustness of detection methods has been considered in prior work on detection of machine generated text \cite{crothers2022adversarial, stiff2022detecting, gagiano2021robustness}.  In one previous work, the robustness of features derived from neural classifiers was compared to robustness of features from statistical classifiers \cite{crothers2022adversarial}.  Unsurprisingly, this work found that incorporating statistical features into feature vectors improved robustness against adversarial attacks that typically target neural classifiers.  Based on these findings, there may be value in leveraging several detection approaches in parallel, necessitating that attackers evade multiple detection models at once.

An often-overlooked element of adversarial attacks against neural text classifiers is the degradation in text quality as a result of adversarial attack. In the text domain, replacing several words using word-level attack such as Textfooler \cite{DBLP:journals/corr/abs-1907-11932} can lead to a result where the meaning of the sentence has changed substantially, or the sentence has been rendered incoherent due to the selection of an ``equivalent" word that does not correctly fit the context.  Character-level attacks that perform character replacements and swaps eventually begin to damage the fluency and credibility of the resulting text \cite{gao2018black}.  A phishing email supposedly sent from a bank, but characterized by unusual word choices and a high frequency of typos, is less desirable to an attacker.  As a result of this, adversarial attacks that fool detection algorithms may fail to fulfill their original purpose in terms of propagating the intended disinformation, or in persuading someone to click a malicious link.  In previous machine text detection research, increased adversarial robustness was accompanied by decreased MAUVE scores in successful attack text \cite{crothers2022adversarial}.  Future applied research might incorporate measures of whether adversarial text that bypasses detection systems would still be effective against targets.


\subsection{Interpretability and Fairness of Detection Methods}

The usage of machine learning models to perform positive detection of machine generated text for the purposes of preventing abuse constitutes a situation where such models are likely to have a negative impact on flagged individuals.  These penalities may range from relatively lightweight (e.g., having to perform a CAPTCHA challenge to post a comment) to more severe (e.g., denial of a scholarship or social media ban).  As a result, as with other automated decision making systems, it is important that such systems operate in a way that is sufficiently fair, transparent, and interpretable. Social or technical research considering potential harms of machine generated text detection is important to ensuring detection systems are ethically acceptable.

The requirement to provide human-understandable explanations has become an important part of trustworthy AI policies, and is reflected in emerging government regulatory guidelines and technology standards related to automated decision making \cite{doi/10.2759/346720, canada_ca_2021, eu_2018_automated, phillips2020four}.  These considerations have also influenced NLG model usage policies \cite{ferrandis_contractor_nguyen_lansky_2022} (discussed further in \S \ref{ssec:trend_pol}).  Early work has been done leveraging random forest models and XGBoost for detection of GPT-2 generated fake reviews, with the goal of providing Shapley Additive Explanations (SHAP) \cite{lundberg2017unified} in machine generated text detection \cite{kowalczyk2022detecting}.  There is a need for future work on methods that are both effective and explainable for machine generated text detection.

Finally, a critical consideration is that certain groups of individuals may be more likely to have their text flagged by machine generated text detection algorithms, either due to characteristics of their writing (such as language background), or due to non-malicious use of translation tools \cite{liang2023gpt}.  For example, it is possible that a detection system designed to prevent a political influence campaign operated using NLG models, may inadvertently end up disproportionately targeting all political speech by individuals who do not natively speak the language of discussion, as has been documented in past research of non-NLG political influence campaigns \cite{crothers2019towards}.  Research that identifies ways to improve detection while maintaining fairness and preventing widespread discrimination is deeply important.

\subsection{Detection Methods Incorporating Human Agency}

As previously mentioned, it is possible that detection of machine generated text may result in suppression of specific individuals or communities in social media whose language background or topics of interest disproportionately cause them to be identified as a false positive by a detection model.  In order to reduce this likelihood, and other ethical harms, the development of machine generated text models that incorporate a human analyst may be of use.  GLTR remains the only tool currently available for detection of machine generated text that explicitly incorporates a human analyst to improve detection \cite{gehrmann-etal-2019-gltr}.  Analysis of GLTR has demonstrated that machine text that fools humans is also more easily detected by algorithms \cite{Ippolito2020AutomaticDO}.  As such, continued development of moderation tools and systems that leave an avenue for human agency and oversight --- guiding principles for trustworthy AI --- is a positive area of future development.  Similar work has already been done in the field of online influence operation research more generally, using Transformer embeddings to chart and cluster social media for free-form exploration by a human analyst \cite{crothers2021mean}, a similar approach may be worthwhile for machine generated text detection.

\subsection{Detection of Abuse Beyond Text Content}


While many of the threat models discussed in \S \ref{sec:threat} can make use of machine generated text detection as part of mitigation strategies, additional methods might be used to facilitate detection outside of text classification.  Work on social bot detection includes additional signals, such as IP addresses and timing of messages, though signals in this domain are also becoming harder to detect over time \cite{stieglitz2017social}.  Chatbot detection can incorporate features derived from human responses \cite{DBLP:journals/corr/abs-2106-01170}.  Prompt injection may bait social bots into exposing themselves \cite{willison_2022}.

On social media platforms, in addition to technical detection approaches, it is likely that many platforms will enact policy approaches to improve user verification as well, providing a greater barrier to entry for fraudulent accounts.  Increased CAPTCHA challenges are already commonplace when platforms are accessed via shared proxy IP addresses, or registered with phone numbers associated with a voice-over-IP (VOIP) services \cite{10.1145/3477142}.  These types of restrictions may become more stringent, with increased user vetting by checking selectors (IP addresses, emails) with third-party reputation services.  While the extent of these measures will vary by platform, it is possible that certain platforms may resort to more stringent verification of an individual's real identity using national IDs or payment methods.  In any case, the asymmetric difficulty of defense versus attack in the current threat environment means that increased scrutiny of new accounts will likely be required to avoid a collapse of trust in online spaces.

\subsection{Defining Model Usage and Disclosure Policies}
\label{ssec:trend_pol}

Undisclosed usage of AI-generated text content is likely to continue to increase, particularly as NLG models are deployed in user-friendly tools such as ChatGPT \cite{openai_2022}, and purpose-built offerings like Jasper \cite{jasper} designed to assist with producing articles and social media content. Increased usage of such tools for generating targeted content may result in situations where individuals online are interacting heavily with content predominantly generated by AI models.


This is cause for concern not just due to the erosion of trustworthy AI principles by not disclosing the usage of AI to a human audience  \cite{doi/10.2759/346720}, but also poses additional ethical problems as NLG models have been found to magnify biases present in training data \cite{solaiman2019release}.   Digital content farms may begin publishing large amounts of predominantly AI-generated text content (articles, blogs, posts, tweets, etc.) and targeting this content towards the audience most likely to engage with it.  Without oversight, this would include highly optimized content that caters to an audience's worst biases and fears --- likely a profitable strategy, as anger and anxiety have a strong link with online virality \cite{berger2012makes}.  Moderation strategies for AI-generated content may include limitations to its use, or notifying readers that they are engaging with AI-generated content to allow them to reconsider how much trust they place in what they are reading.



Usage and disclosure policies for online platforms are a worthwhile area of future development, whether those take the role of AI usage restrictions (such as those related to generative art \cite{benj_ai_ban}), or mandated public disclosure of AI-generated content.  Model publishers can also influence behaviour of law-abiding entities by adjusting the licenses of released models to mandate disclosure.  AI model BLOOM was released under the first version of the Responsible AI License (RAIL) \cite{ferrandis_contractor_nguyen_lansky_2022}.  The conditions of this license include a requirement for disclosure, an explicit ban on malicious abuse, and a prohibition of specific use-cases (including automated decision making with a potential negative impact, which aligns with regulation terminology in the EU \cite{eu_2018_automated} and Canada \cite{canada_ca_2021}).  An effective combination of usage policies and AI software licenses may improve the ethical rigour in how powerful NLG models are used in practice, though great care must be taken in crafting such restrictions.
\section{Conclusion}
\label{sec:conclusion}

In this survey, we provided a comprehensive overview of detection methods for machine generated text, carefully evaluating the technical and social benefits of different approaches and including novel research focusing on topics such as adversarial robustness and explainability.  We provided context for this review with an overview of natural language generation (NLG) models, and a deep analysis of current threat models.  Our exploration of threat models, when viewed alongside our survey on applied detection research, suggests that current domain-specific defenses are not adequate to defend against the vast majority of upcoming threat models. Recent NLG advances, which combine dramatic improvements in text quality with unparalleled ease-of-use, further highlight the urgent nature of developing improved defenses against abuse of machine generated text.

Our central conclusion is that the field of machine generated text detection has a multitude of open problems that urgently need attention in order to provide suitable defenses against widely-available NLG models.  Existing detection methodologies often do not reflect realistic settings of class imbalance or unknown model architectures, nor do they incorporate sufficient transparency and fairness methods to ensure that such detection systems will not themselves cause harm. Preventing widespread harms from NLG models will require coordinated effort across technical and social domains, necessitating alignment between AI researchers, cybersecurity professionals, and non-technical experts.  While there are a wide range of threat models and open research problems to consider, tackling these challenges is essential for humans to realize the benefits of high-capacity NLG systems while reducing the damage caused by their inevitable abuse.

\bibliographystyle{acm-bib/ACM-Reference-Format}
\bibliography{ref}

\end{document}